\documentclass[11pt,a4paper]{article}
\RequirePackage{graphicx,xcolor,ae,fancyvrb}
\RequirePackage[T1]{fontenc}
\usepackage{thumbpdf,lmodern}
\usepackage{amsmath}
\usepackage{amssymb}
\usepackage{dsfont}
\usepackage{algorithm}
\usepackage{algpseudocode}
\usepackage{cleveref}
\usepackage{booktabs}
\usepackage{multirow}
\usepackage{graphicx}
\usepackage{adjustbox}
\usepackage{natbib}
\usepackage{url}
\DefineVerbatimEnvironment{CodeInput}{Verbatim}{fontshape=sl}
\let\proglang=\textsf
\let\code=\texttt
\newcommand{\class}[1]{`\code{#1}'}
\newcommand{\fct}[1]{\code{#1()}}
\newcommand{\tr}{^{\text{\tiny T}}}
\newcommand{\dis}{\mbox{d}}

\newcommand{\diag}{\mbox{diag}}
\newcommand{\argmax}{\mbox{argmax}}
\newcommand{\Ker}{K}
\newcommand{\Prob}{\mathsf{P}}
\newcommand{\pkg}[1]{{\fontseries{m}\fontseries{b}\selectfont #1}}

\begin{document}
\def\thefootnote{$\dagger$}\footnotetext{These authors contributed equally to this work.}\def\thefootnote{\arabic{footnote}}
\def\thefootnote{*}\footnotetext{Corresponding author.}\def\thefootnote{\arabic{footnote}}  
\title{\pkg{SurvLIMEpy}: A \proglang{Python} package implementing SurvLIME}
\author{
	 Cristian Pachón-García$^\dagger$\\
	 \small{Universitat Politècnica de Catalunya}\\
	\and
	Carlos Hernández-Pérez$^\dagger$\\
	 \small{Universitat Politècnica de Catalunya}\\
	\and
	Pedro Delicado\\
	 \small{Universitat Politècnica de Catalunya}\\
	\and
	Verónica Vilaplana*\\
	 \small{Universitat Politècnica de Catalunya}\\
}

\date{\today}
\maketitle

\begin{abstract}
In this paper we present \pkg{SurvLIMEpy}, an open-source \proglang{Python} package that implements the SurvLIME algorithm. This method allows to compute local feature importance for machine learning algorithms designed for modelling Survival Analysis data. Our implementation takes advantage of the parallelisation paradigm as all computations are performed in a matrix-wise fashion which speeds up execution time. Additionally, \pkg{SurvLIMEpy} assists the user with visualization tools to better understand the result of the algorithm. The package supports a wide variety of survival models, from the Cox Proportional Hazards Model to deep learning models such as DeepHit or DeepSurv. Two types of experiments are presented in this paper. First, by means of simulated data, we study the ability of the algorithm to capture the importance of the features. Second, we use three open source survival datasets together with a set of survival algorithms in order to demonstrate how \pkg{SurvLIMEpy} behaves when applied to different models.
	
	  \proglang{Python}.
\end{abstract}
\section{Introduction}
\label{sec:introduction}
Survival Analysis, also known as time-to-event analysis, is a field of Statistics that aims to study the time until a certain event of interest occurs. The reference approach for modelling the survival time is the Cox Proportional Hazards Model \citep{CoxAlgorithm}.

A survival study follows up a set of individuals among whom some will eventually experience the event of interest. Due to the nature of these studies, it is common to find the problem of censorship. An event may not be observed in all individuals due to lost to follow-up, dropping from the study or finishing the study without the event occurring. The Cox Proportional Hazards Model takes into account the phenomenon of censorship, since  the estimation of the parameters is done through a likelihood function that deals with censorship.

Nowadays, a wide set of machine learning models are able to tackle Survival Analysis problems. Among them, it is worth highlighting Random Survival Forest \citep{random_survival_forest}, survival regression with accelerated failure time model in XGBoost \citep{xgboost_survival} or adaptations of deep learning algorithms for Survival Analysis such as DeepHit \citep{deepHit} or DeepSurv \citep{deepSurv}. These models have proven to have good prediction capacity, as reported in \cite{SurvivalSurvey, SurvivalSurvey_2, DeepLearningSurvival}.

Despite the continuous advances in the development of machine learning algorithms for healthcare applications, their adoption by medical practitioners and policy makers in public health is still limited. One of the main reasons is the black-box nature of most of the models, in the sense that the reasoning behind their predictions is often hidden from the user. 

Interpretable Machine Learning (or, equivalently, eXplainable Artificial Intelligence, XAI for short) is a recent field born out of the need to derive explanations from machine learning models \citep{xai_need}. Two popular interpretability methods are LIME \citep{lime} and SHAP \citep{shap}, which provide explanations locally around a test example. Although they are extensively used \citep{using_lime}, these algorithms are not designed to deal with time-to-event data, which invalidates their use in survival applications.

The SurvLIME algorithm \citep{survlime}, inspired by LIME, was the first method presented in order to interpret black box survival models. This method aims to compute local interpretability by means of providing a ranking among the set of features for a given individual $\mathbf{x_{*}}$, but unlike the methods mentioned previously, it considers the time space to provide explanations. First, it generates a set of neighbours, then it obtains a set of predictions for the neighbours and, finally, a Cox Proportional Hazards Model (local explainer) is fitted, minimising the distance between the predictions provided by the black box model and the predictions provided by the local explainer. The coefficients of the local model serve as an explanation for the survival machine learning model. 

In a recent work, it was presented SurvSHAP(t) \citep{survshap_t}, an interpretability method inspired by SHAP algorithm designed to explain time-to-event machine learning models. In short, the explanation is provided by means of a time-dependent function. The time space is included in the explanation with the goal of detecting possible dependencies between the features and the time. Alongside this method, implementations of SurvLIME and SurvSHAP(t) algorithms were presented in the \proglang{R} package \pkg{survexp} \citep{survexp}.

In this work we present an open-sourced \proglang{Python} package, \pkg{SurvLIMEpy}, which implements the SurvLIME algorithm. The package offers some degrees of freedom to the users. For instance, they can choose how to obtain the neighbours of the test individual, the distance metric to be minimised or to carry out a Monte-Carlo simulation. Furthermore, we provide details on how to use it, illustrated with some open source survival datasets as well as a simulation study, aiming to analyse the performance of the SurvLIME algorithm. As far as we know, this is the first \proglang{Python} implementation of this method.

The rest of the paper is organised as follows: in Section \ref{sec:survlime_algorithm}, the most relevant parts of the SurvLIME algorithm are presented. In Section \ref{sec:package_implementation}, we introduce the package implementation. Additionally, a use case is provided. In Section \ref{sec:experiments}, we present some experiments conducted with both simulated and real datasets. In this section, we use some of the state-of-the-art machine learning and deep learning algorithms for Survival Analysis in order to show how \pkg{SurvLIMEpy} is used with those models. Finally, conclusions are presented in Section \ref{sec:conclusions}. 

\section{SurvLIME algorithm}
\label{sec:survlime_algorithm}
In this section we summarise the SurvLIME algorithm, which was presented in \cite{survlime}. We first introduce some notation. Let $\mathbf{D}=\{(\mathbf{x_j}, \tau_j, \delta_j)\}$ $j \in \{1, \dots, n\}$ be a dataset of triplets that represent individuals, where $\mathbf{x_j} \in \mathbb{R}^p$ is a $p$-dimensional feature vector, $\tau_j$ is the time to event or lost to follow-up time, and $\delta_j$ is the event indicator (1 means the event occurs and 0 otherwise). Let $t_1 < \cdots < t_{m+1}$ be the distinct times from $\{\tau_1, \dots, \tau_n\}$. Let
\begin{align*}
  \hat{H}  \colon \mathbb{R}^p \times \mathbb{R}_{>0}  &\to \mathbb{R}_{>0} \\
  (\mathbf{x}, t) &\mapsto \hat{H}(\mathbf{x}, t)
\end{align*}
be the already trained machine learning model that predicts the Cumulative Hazard Function (CHF; see Appendix \ref{appendix:surv-analysis} for more details) of an individual $\mathbf{x}$ at time $t$. In SurvLIME \citep{survlime}, the authors prove that $\hat{H}(\mathbf{x}, t)$ can be written as
\begin{equation}
    \label{eq:chf_piecewise}
    \hat{H}(\mathbf{x}, t) = \sum_{i=1}^{m+1}\hat{H}_i(\mathbf{x})\mathds{1}_{\Omega_i}(t),
\end{equation}
where $\Omega_i = [t_i, t_{i+1})$, being $t_{m+2} = t_{m+1} + \gamma$ ($\gamma$ a small positive number) and $\mathds{1}_{\Omega_i}(t)$ the indicator function (1 if $t \in \Omega_i$ and 0 otherwise). In the original paper, the authors did not specify any value for $\gamma$. In our implementation, we use $10^{-6}$.

It is important to note that the function $\hat{H}_i(\mathbf{x})$ is constant in $\Omega_i$. Therefore, if $\Omega=\cup_{i=1}^{m+1}\Omega_j$ and $g \colon \Omega \to \mathbb{R}$ is a monotone function, then
\begin{equation}
    \label{eq:monotone_piecewise}
    g(\hat{H}(\mathbf{x}, t)) = \sum_{i=1}^{m+1}g\left[\hat{H}_i(\mathbf{x})\right]\mathds{1}_{\Omega_i}(t).
\end{equation}

Given the prediction provided by the black-box model
$\hat{H}(\mathbf{x}_*, t)$ for an individual $\mathbf{x}_*$, SurvLIME finds the importance of each feature by means of approximating $\hat{H}(\mathbf{x}_*, t)$ by the Cox Proportional Hazards Model,
$\hat{H}_{cox}(\mathbf{x}_*, t)=H_0(t)\exp(\hat{\boldsymbol\beta}\tr\mathbf{x}_*)$ (see Appendix \ref{appendix:surv-analysis} for more details). 

Applying Expression (\ref{eq:chf_piecewise}) to the 
Cox Proportional Hazards Model, a new expression for this model is obtained:
\begin{equation}
    \label{eq:piecewise_cox}
    \hat{H}_{cox}(\mathbf{x}_*, t) = \sum_{i=1}^{m+1}\left[\hat{H}_{0}(t_i)\exp({\hat{\boldsymbol\beta}\tr\mathbf{x}_*})\right]\mathds{1}_{\Omega_i}(t).
\end{equation}

After fixing the individual $\mathbf{x}_*$, both functions, $\hat{H}(\mathbf{x}_*, t)$ and $\hat{H}_{cox}(\mathbf{x}_*, t)$, only depend on t. Taking the logarithms $\phi(t) = \ln[\hat{H}(\mathbf{x}_*, t)]$ and
$\phi_{cox}(t) = \ln[\hat{H}_{cox}(\mathbf{x}_*, t)]$ and taking into account Expression (\ref{eq:monotone_piecewise}), the following can be derived:
\begin{equation}
    \label{eq:ln_bb_model}
    \phi(t)=\sum_{i=1}^{m+1} \ln\left[\hat{H}_i(\mathbf{x}_*)\right]\mathds{1}_{\Omega_i}(t),
\end{equation}

\begin{equation}
    \label{eq:ln_cox_model}
    \phi_{cox}(t)=\sum_{i=1}^{m+1} \left(\ln\left[\hat{H}_{0}(t_i)\right] + \hat{\boldsymbol\beta}\tr\mathbf{x}_*\right)\mathds{1}_{\Omega_i}(t).
\end{equation}

Let us consider $\alpha(t) = \phi(t)-\phi_{cox}(t) = \sum_{i=1}^{m+1}(\ln[\hat{H}_i(\mathbf{x}_*)] - \ln[\hat{H}_{0}(t_i)] - \hat{\boldsymbol\beta}\tr\mathbf{x}_*) \mathds{1}_{\Omega_i}(t)$. Since $\alpha(t)$ is a piecewise constant function and $s(t) = t^2$ is a monotone function for $t \geq 0$, we can use Expression (\ref{eq:monotone_piecewise}) to write $(\phi - \phi_{cox})^2$ as a piecewise constant function,
\begin{equation}
    \begin{aligned}
        \label{eq:square_difference}
        s(\alpha(t)) = \left(\phi(t) - \phi_{cox}(t)\right)^2 &= \sum_{i=1}^{m+1}\left(\ln\left[\hat{H}_i(\mathbf{x}_*)\right] - \ln\left[\hat{H}_{0}(t_i)\right] - \hat{\boldsymbol\beta}\tr\mathbf{x}_*\right)^2 \mathds{1}_{\Omega_i}(t).
    \end{aligned}
\end{equation}

\newpage
The next step is to find a vector $\hat{\boldsymbol\beta}$ that minimises the $\ell^2$ distance between $\phi$ and $\phi_{cox}$. Taking into account that both functions are considered in $\Omega$,
\begin{equation}
    \begin{aligned}
        \label{eq:distance}
        \dis^2(\phi, \phi_{cox}) &= \lVert \phi -  \phi_{cox} \rVert_{2}^2\\
        &= \int_{\Omega} \left[\phi(t) - \phi_{cox}(t)\right]^2 \, dt\\
        &= \int_{\Omega} \sum_{i=1}^{m+1}\left(\ln\left[\hat{H}_i(\mathbf{x}_*)\right] - \ln\left[\hat{H}_{0}(t_i)\right] - \hat{\boldsymbol\beta}\tr\mathbf{x}_*\right)^2 \mathds{1}_{\Omega_i}(t) \, dt \\
        &= \sum_{i=1}^{m+1} \left( \ln\left[\hat{H}_i(\mathbf{x}_*)\right] - \ln\left[\hat{H}_{0}(t_i)\right] - \hat{\boldsymbol\beta}\tr\mathbf{x}_* \right)^2 \Delta t_i, 
    \end{aligned}
\end{equation}
where $\Delta t_i = (t_{i+1} - t_i)$. We have used Expression (\ref{eq:square_difference}) and that $\ln[\hat{H}_i(\mathbf{x}_*)] - \ln[\hat{H}_{0}(t_i)] - \hat{\boldsymbol\beta}\tr\mathbf{x}_*$ does not depend on $t$ to derive the previous expression.

Since SurvLIME is inspired by LIME, a set of $N$ points $\{\mathbf{e}_1, \dots, \mathbf{e}_N\}$ are generated in a neighbourhood of $\mathbf{x_*}$, and the objective function is expressed in terms of these points and their corresponding weights, that depend on the distance between the $N$ points and the individual $\mathbf{x_*}$. Applying Expression (\ref{eq:distance}) for all the neighbours $\mathbf{e}_k$, the following objective is obtained:
\begin{equation}
    \label{eq:neighbours}
    \min_{\hat{\boldsymbol\beta}}\sum_{k=1}^N w_k \sum_{i=1}^{m+1} \left( \ln\left[\hat{H}_i(\mathbf{e}_k)\right] - \ln\left[\hat{H}_{0}(t_i)\right] - \hat{\boldsymbol\beta}\tr\mathbf{e}_k \right)^2 \Delta t_i.
\end{equation}

For each point $\mathbf{e}_k$ a weight is computed using a kernel function, $w_k = \Ker(\mathbf{x}_*, \mathbf{e}_k)$; the closer $\mathbf{e}_k$ is to $\mathbf{x}_*$, the higher the value of $w_k$ is. 

Finally, the authors introduce weights
$u_{ki} = \hat{H}_i(\mathbf{e}_k)/\ln(\hat{H}_i(\mathbf{e}_k))$ in Expression (\ref{eq:neighbours}), as the difference between $\hat{H}(\mathbf{x}, t)$ and $\hat{H}_{cox}(\mathbf{x}, t)$ could be significantly different from the distance between their logarithms. Therefore, the goal is to minimise the following expression:
\begin{equation}
    \label{eq:minimise}
    \min_{\hat{\boldsymbol\beta}}\sum_{k=1}^N w_k \sum_{i=1}^{m+1} u_{ki}^2 \left( \ln\left[\hat{H}_i(\mathbf{e}_k)\right] - \ln\left[\hat{H}_{0}(t_i)\right] - \hat{\boldsymbol\beta}\tr\mathbf{e}_k \right)^2 \Delta t_i.
\end{equation}

Note that the first two factors in Expression (\ref{eq:minimise}) are quadratic and the last one is positive. Thus, the product is a convex function. Since we are considering a weighted sum of convex functions, the resulting expression is also convex. Therefore, there exists a solution for this problem.

Algorithm \ref{alg:survlime} summarises how to proceed in order to obtain the coefficients $\hat{\boldsymbol\beta}$ for the local Cox Proportional Hazards Model approximation. In case all the features are standardised, feature $i$ is more important than feature $j$ if $\lvert \hat{\beta}_i \rvert > \lvert \hat{\beta}_j \rvert$. If they are not standardised, $\mathbf{x}_*$ must be taken into account to perform the comparison: feature $i$ is more important than feature $j$ if $\lvert \hat{\beta}_i x_{*i} \rvert > \lvert \hat{\beta}_j x_{*j} \rvert$.

\begin{algorithm}[!h]
    \caption{SurvLIME algorithm.}
    \label{alg:survlime}
    \begin{algorithmic}
        \Require Input variables:
        \vspace{-2mm}
        \begin{itemize}
            \item Training dataset $\mathbf{D}=\{(\mathbf{x_j}, \tau_j, \delta_j)\}$, $j \in \{1, \dots, n\}$.
            \item Individual of interest $\mathbf{x_*}$.
            \item Number of neighbours to generate $N$.
            \item Black-box model for the Cumulative Hazard Function 
            $\hat{H} \colon \mathbb{R}^p \times \mathbb{R}_{>0}  \to \mathbb{R}_{>0}$.
            \item Kernel function 
            $\Ker \colon \mathbb{R}^p \times \mathbb{R}^p \to \mathbb{R}_{>0}$ to compute the weights according to the distance to $\mathbf{x_*}$.
        \end{itemize}
        \Ensure Obtain vector $\hat{\boldsymbol\beta}$ for the local Cox Proportional Hazards Model approximation.
        \begin{enumerate}
            \item Obtain the distinct times $t_i$, $i \in \{1, \dots, m+1\}$ from $\mathbf{D}$.
            \item Estimate the baseline Cumulative Hazard Function $\hat{H}_0(t)$ using $\mathbf{D}$ and the Nelson-Aalen estimator.
            \item Generate $N$ neighbours of $\mathbf{x_*}$: $\{\mathbf{e}_1, \dots, \mathbf{e}_N\}$.
            \item For each time step $t_i$ and for each $\mathbf{e}_k$, obtain the prediction $\hat{H}_i(\mathbf{e}_k)$.
            \item Obtain $\ln\left(\hat{H}_i(\mathbf{e}_k)\right)$.
            \item For each $\mathbf{e}_k$, obtain the weight $w_k=\Ker(\mathbf{x}_*, \mathbf{e}_k)$.
            \item For each time step $t_i$ and for each $\mathbf{e}_k$, obtain $u_{ki} = \hat{H}_i(\mathbf{e}_k)/\ln\left(\hat{H}_i(\mathbf{e}_k)\right)$.
            \item Solve the convex optimisation problem stated in Expression (\ref{eq:minimise}).
        \end{enumerate}
    \end{algorithmic}
\end{algorithm}

\section{Package implementation}
\label{sec:package_implementation}
In this section, we introduce \pkg{SurvLIMEpy}, an open-source \proglang{Python} package that implements the SurvLIME algorithm. It is stored in the \proglang{Python} Package Index (\proglang{PyPI})\footnote{\url{https://pypi.org/project/survlimepy/}} and the source code is available at \proglang{GitHub}\footnote{\url{https://github.com/imatge-upc/SurvLIMEpy}}. Additionally, we present a detailed explanation of the implementation as well as some additional flexibility provided to the package.

Section \ref{sec:matrix_expresion} introduces a matrix-wise formulation for Expression (\ref{eq:minimise}). \Cref{sec:neighbour_kernel,sec:functional_norm,sec:survival_models}  describe the parts of the package that the user can adjust. \Cref{sec:code,sec:example} describe how to use the package and some code examples are given.

\subsection{Matrix-wise formulation}
\label{sec:matrix_expresion}
In order to apply a parallelism paradigm, and thus reduce the total execution time, the optimisation problem can be formulated matrix-wise. Before developing it, we introduce some notation. Let $\mathbf{1}_d$ be the vector of ones of size $d$, i.e., $\mathbf{1}_d = (1, \dots, 1)\tr$. Let $\mathbf{A}=(a_{ij})$ and $\mathbf{C}=(c_{ij})$ be two matrices of the same size. By $\oslash$ we denote the element-wise division between $\mathbf{A}$ and $\mathbf{C}$. Likewise, $\odot$ denotes the element-wise product.

The first step is to find the matrix expression for $\ln[\hat{H}_0(t_i)]$. Let $\mathbf{v}_0$ be the component-wise logarithm of the baseline Cumulative Hazard function evaluated at each distinct time, i.e.,
$\mathbf{v}_0 = (\ln[\hat{H}_0(t_1)], \dots, \ln[\hat{H}_0(t_{m+1})])\tr$. To produce a matrix, let us consider the product between $\mathbf{1}_N$ and $\mathbf{v}_0\tr$,
$\mathbf{L}_0=\mathbf{1}_N\mathbf{v}_0\tr$.  $\mathbf{L}_0$ is a matrix of size $N \times(m+1)$. Note that all the rows contain exactly the same vector $\mathbf{v}_0$.

After that, a matrix $\mathbf{E}$ containing $N$ neighbours is obtained. The size of $\mathbf{E}$ is $N \times p$. Each row in $\mathbf{E}$, denoted by $\mathbf{e}_k$, is a neighbour of $\mathbf{x}_*$. To find the matrix expression for $\ln[\hat{H}_i(\mathbf{e}_k)]$, let $\mathbf{V} = (v_{ki})$ be the matrix that contains the values of the  Cumulative Hazard Function for the neighbours evaluated at each distinct time, i.e., $v_{ki} = \hat{H}_i(\mathbf{e}_k)$. Let us consider the component-wise logarithm of $\mathbf{V}$, $\mathbf{L} = (\ln[v_{ki}])$. $\mathbf{V}$ and $\mathbf{L}$ are of size $N\times(m+1)$.

Next, we find the matrix-wise expression for $[\hat{H}_i(\mathbf{e}_k)/\ln(\hat{H}_i(\mathbf{e}_k)]^2$. Let $\mathbf{M}$ be the resulting matrix of the element-wise division between $\mathbf{V}$ and $\mathbf{L}$, i.e.,
$\mathbf{M} = \mathbf{V} \oslash \mathbf{L}$ and let 
$\mathbf{M}_2 = \mathbf{M} \odot \mathbf{M}$, which is of size $N \times (m+1)$.

The next step is to find the matrix expression for $\hat{\boldsymbol\beta}\tr\mathbf{e}_k$. Let $\hat{\boldsymbol\beta}$ be the unknown vector (of size $p$) we are looking for. Let us consider the product between $\mathbf{E}$ and $\hat{\boldsymbol\beta}$, $\mathbf{\tilde{p}}=\mathbf{E}\hat{\boldsymbol\beta}$, which is a vector of size $N$ and whose $k-th$ component is $\hat{\boldsymbol\beta}\tr\mathbf{e}_k$. To obtain a matrix of size $N\times(m+1)$, let us consider the product between $\mathbf{\tilde{p}}$ and $\mathbf{1}_{m+1}$,
$\boldsymbol\Lambda = \mathbf{\tilde{p}}\mathbf{1}_{m+1}\tr$. All the columns in $\boldsymbol\Lambda$ contain the same vector  $\mathbf{\tilde{p}}$.

Let us obtain the matrix-wise expression of
$(\ln[\hat{H}_i(\mathbf{e}_k)] - \ln[\hat{H}_{0}(t_i)] - \hat{\boldsymbol\beta}\tr\mathbf{e}_k)^2$. First, let us consider
$\boldsymbol\Theta = \mathbf{L} - \mathbf{L}_0 - \boldsymbol\Lambda$. Note that the size of the matrix $\boldsymbol\Theta$ is $N \times (m+1)$. Second, let us consider the element-wise square of the previous matrix, denoted by $\boldsymbol\Theta_2$, i.e., $\boldsymbol\Theta_2 = \boldsymbol\Theta \odot \boldsymbol\Theta$. The component $(k, i)$ of the previous matrix contains the desired expression, where $k \in \{1, \dots, N\}$ and $i \in \{1, \dots, m+1\}$.

Now, we obtain the matrix expression for $u_{ki}^2(\ln[\hat{H}_i(\mathbf{e}_k)] - \ln[\hat{H}_{0}(t_i)] - \hat{\boldsymbol\beta}\tr\mathbf{e}_k)^2$. To do that, let $\boldsymbol\Pi$ be the matrix obtained by the element-wise multiplication between $\mathbf{M}_2$ and $\boldsymbol\Theta_2$,
$\boldsymbol\Pi = \mathbf{M}_2 \odot \boldsymbol\Theta_2$. $\boldsymbol\Pi$ is of size $N \times (m+1)$.

Let $\mathbf{t}$ be the vector of size $m+2$ containing the distinct times (we apply the same consideration as in Section \ref{sec:survlime_algorithm}, i.e., $t_{m+2} = t_{m+1} + \gamma$). Let 
$\boldsymbol\psi_t$ be the vector of time differences between two consecutive distinct times, i.e., $\boldsymbol\psi_t = (t_2 - t_1, \dots, t_{m+2} - t_{m+1})\tr$, which is a vector of size $m+1$.

To obtain $\sum_{i=1}^{m+1}u_{ki}^2(\ln[\hat{H}_i(\mathbf{e}_k)] - \ln[\hat{H}_{0}(t_i)] - \hat{\boldsymbol\beta}\tr\mathbf{e}_k)^2\Delta t_i$ matrix-wise, let $\boldsymbol\pi$ be the resulting vector of multiplying the matrix $\boldsymbol\Pi$ and the vector $\psi_t$, i.e.,
$\boldsymbol\pi = \boldsymbol\Pi\boldsymbol\psi_t$. The vector $\boldsymbol\pi$  is of size $N$ and the $k-th$ component of it contains the desired expression.

Finally, let $\mathbf{w}$ be the vector of weights for the neighbours, which is of size $N$. This vector is obtained by applying the kernel function over all the neighbours, i.e., $w_k=\Ker(\mathbf{x}_*, \mathbf{e}_k)$. Then, Expression (\ref{eq:minimise}) can be formulated as
\begin{equation}
    \label{eq:minimise_matrix}
    \min_{\hat{\boldsymbol\beta}}\mathbf{w}\tr \boldsymbol\pi,
\end{equation}
where the vector $\boldsymbol\pi$ depends on the vector $\hat{\boldsymbol\beta}$. Algorithm \ref{alg:survlime_matrix} summarises this matrix-wise implementation.

In order to find a numerical solution for Expression (\ref{eq:minimise_matrix}), we use the \pkg{cvxpy} package \citep{cvxpy}. This library contains functionalities that allow to perform matrix-wise operations as well as element-wise operations. Moreover, \pkg{cvxpy} library allows the user to choose the solver applied to the optimisation algorithm. In our implementation, we use the default option, which is the Operator Splitting Quadratic Program solver, OSQP for short \citep{osqp}. 

\begin{algorithm}[!h]
    \caption{Matrix-wise SurvLIME algorithm.}
    \label{alg:survlime_matrix}
    \begin{algorithmic}
        \Require Input variables:
        \vspace{-2mm}
        \begin{itemize}
            \item Training dataset $\mathbf{D}=\{(\mathbf{x_j}, \tau_j, \delta_j)\}$, $j \in \{1, \dots, n\}$.
            \item Individual of interest $\mathbf{x_*}$.
            \item Number of neighbours to generate $N$.
            \item Black-box model for the Cumulative Hazard Function 
            $\hat{H} \colon \mathbb{R}^p \times \mathbb{R}_{>0}  \to \mathbb{R}_{>0}$.
            \item Kernel function 
            $\Ker \colon \mathbb{R}^p \times \mathbb{R}^p \to \mathbb{R}_{>0}$ to compute the weights according to the distance to $\mathbf{x_*}$.
        \end{itemize}
        \Ensure Obtain vector $\hat{\boldsymbol\beta}$ for the local Cox Proportional Hazards Model approximation.
        \begin{enumerate}
            \item Obtain the vector of distinct times $\mathbf{t}$ from $\mathbf{D}$.
            \item Obtain the vector $\mathbf{v}_0$, the component-wise logarithm of baseline Cumulative Hazard Function evaluated at each distinct time $t_i$ using $\mathbf{D}$ and the Nelson-Aalen estimator.
            \item Generate matrix of neighbours $\mathbf{E}$.
            \item Obtain matrix $\mathbf{V}=\hat{H}(\mathbf{E}, \mathbf{t})$.
            \item Calculate component-wise logarithm of  $\mathbf{V}$, $\mathbf{L} = \ln(\mathbf{V})$.
            \item Obtain the vector of weights, $\mathbf{w}=\Ker(\mathbf{x}_*, \mathbf{E})$.
            \item Obtain the matrix of weights $\mathbf{M}_2$.
            \item Solve the convex optimisation problem stated in Expression (\ref{eq:minimise_matrix}).
        \end{enumerate}
    \end{algorithmic}
\end{algorithm}

\subsection{Neighbour generation and kernel function}
\label{sec:neighbour_kernel}
The neighbour generating process is not specified in the original LIME paper nor in the SurvLIME publication. 
As reported in \cite{molnar_book}, this issue requires great care since explanations provided by the algorithm may vary depending on how the neighbours are generated.

%The generation of neighbors can be done in multiple ways. 
In our implementation, we use a non-parametric kernel density estimation approach. 
Let $\mathbf{x}_1, \dots, \mathbf{x}_n$, a $p$ dimensional sample drawn from a random variable $\mathcal{X}$ with density function $f$. 
Let $\hat{\sigma}_j$ be the sampling standard deviation of the the $j-th$ component of $\mathcal{X}$.
For a point $\mathbf{x}\in \mathbb{R}^p$, a kernel-type estimator of $f(\mathbf{x})$ is
\[
\hat{f}(\mathbf{x}) = \frac{1}{n b^p \prod_{j=1}^p \hat{\sigma}_j}
\sum_{i=1}^n \exp\left(-\frac{1}{2 b^2} \lVert \mathbf{x}- \mathbf{x}_i\rVert_{s}^2\right),
\]
where $\lVert \mathbf{x}- \mathbf{x}_i\rVert_{s}= 
\sqrt{\sum_{j=1}^p (x_j-x_{ij})^2/\hat{\sigma}^2_j}$ is the Euclidean distance between the standardised versions of $\mathbf{x}$ and $\mathbf{x}_i$,
and $b$ is the bandwidth, a tuning parameter which, by default, we fix as 
$b=[4/(n[p+2])]^{1/(p+4)}$, following the Normal reference rule \cite[page 87]{silverman}.
Observe that $\hat{f}(\mathbf{x})$ is a mixture of $n$ multivariate Normal density functions, each with weight $1/n$, mean value $\mathbf{x}_i$ and common covariance matrix
$\boldsymbol{\hat{\Sigma}}=b^2 \cdot \diag(\hat{\sigma}_1^2, \dots, \hat{\sigma}_p^2)$.
We consider such a Normal distribution centering it at a point of interest $\mathbf{x}_*$: $\mathcal{N}(\mathbf{x}_*,\boldsymbol{\hat{\Sigma}})$. 

First, a matrix containing a set of $N$ neighbours, denoted by $\mathbf{E}$, is generated, each row $\mathbf{e}_k$ coming from a $\mathcal{N}(\mathbf{x}_*,\boldsymbol{\hat{\Sigma}})$, $k\in\{1,\dots,N\}$.
Afterwards, the weight $w_k$ of neighbour $\mathbf{e}_k$ is computed as the value of the density function of the $\mathcal{N}(\mathbf{x}_*,\boldsymbol{\hat{\Sigma}})$ evaluated at $\mathbf{e}_k$. 

\subsection{Functional norm}
\label{sec:functional_norm}
While the original publication uses the $\ell^2$ functional norm to measure the distance between $\phi$ and $\phi_{cox}$ in Expression (\ref{eq:distance}), other works such as SurvLIME-Inf \citep{SurvlimeInf} use $\ell^{\infty}$. The authors of  SurvLIME-Inf claim that this norm speeds up the execution time when solving the optimisation problem.

In our implementation, the computational gain of using the infinity norm was negligible when solving the problem in a matrix-wise formulation as explained in Section \ref{sec:matrix_expresion}. The $\ell^2$ is set as the default distance in our implementation. However, the user can choose other norms.

\subsection{Supported survival models}
\label{sec:survival_models}
Throughout this work, we represent a survival model as a function 
$\hat{H}  \colon \mathbb{R}^p \times \mathbb{R}_{>0}  \to \mathbb{R}_{>0}$. However, the packages that implement the different models do not work in the same way, since their implementations employ a function that takes as input a vector of size $p$ and outputs a vector of size $m+1$, where $m+1$ is the number of distinct times (see Section \ref{sec:survlime_algorithm} for more details). Therefore, the output is a vector containing the Cumulative Hazard Function evaluated at each distinct time, i.e., $\hat{H}  \colon \mathbb{R}^p \to \mathbb{R}_{>0}^{m+1}$. 

Our package can manage multiple types of survival models. In addition to the Cox Proportional Hazards Model \citep{CoxAlgorithm}, which is implemented in the \pkg{sksurv} library \citep{sksurv}, \pkg{SurvLIMEpy} also manages other algorithms: Random Survival Forest \citep{random_survival_forest}, implemented in the \pkg{sksurv} library, Survival regression with accelerated failure time model in XGBoost \citep{xgboost_survival}, implemented in the \pkg{xgbse} library \citep{xgbse}, DeepHit \citep{deepHit} and DeepSurv \citep{deepSurv}, both implemented in the \pkg{pycox} library \citep{pycox}.

The set of times for which the models compute a prediction can differ across models and their implementations. Whereas Cox Proportional Hazards Model, Random Survival Forest and DeepSurv offer a prediction for each distinct time, $\{t_1, \dots, t_{m+1}\}$, 
the other models work differently: for a given integer $q+1$, they estimate the $q+1$ quantiles $\{\tilde{t}_1, \dots, \tilde{t}_{q+1}\}$ and then, they offer a prediction for each $\tilde{t}_j$.

The first models output a vector of size $m+1$, $(\hat{H}(t_1), \dots, \hat{H}(t_{m+1}))\tr$. The second models output a vector of size $q+1$, $(\hat{H}(\tilde{t}_1), \dots, \hat{H}(\tilde{t}_{q+1}))\tr$. Since SurvLIME requires the output of the model to be a vector of length $m+1$, we use linear interpolation in order to fulfill this condition. All of the machine learning packages provide a variable specifying the set of times for which the model provides a prediction. We use this variable to perform the interpolation.

We choose to ensure the integration of the aforementioned machine learning algorithms with \pkg{SurvLIMEpy} as they are the most predominant in the field \citep{SurvivalSurvey, SurvivalSurvey_2, DeepLearningSurvival}. In \Cref{sec:code,sec:example} there are more details on how to provide the prediction function to the package. Note that if a new survival package is developed, \pkg{SurvLIMEpy} will support it as long as the output provided by the predict function is a vector of length $q+1$, $0< q \leq m$.

Usually, the libraries designed to create machine learning algorithms for survival analysis make available two functions to create predictions, one for the Cumulative Hazard Function (CHF) and another one for the Survival Function (SF). For example, for the \pkg{sksurv} package this functions are
\code{predict\_cumulative\_hazard\_function} and \code{predict\_survival\_function}, respectively. \pkg{SurvLIMEpy} has been developed to work with both of them. The user should specify which prediction function is using. By default, the package assumes that the prediction function is for the CHF. In case of working with the SF, a set of transformations is performed in order to work with CHF (see Appendix \ref{appendix:surv-analysis}, where the relationship between the CHF and the SF is explained).

\subsection{Package structure}
\label{sec:code}
The class \class{SurvLimeExplainer} is used as the main object of the package to computes feature importance.
\begin{CodeInput}
    SurvLimeExplainer(
        training_features, training_events, training_times,
        model_output_times, H0, kernel_width,
        functional_norm, random_state
    )
\end{CodeInput}
\begin{itemize}
    \item \code{training\_features}: Matrix of features of size $n \times p$, where $n$ is the number of individuals and $p$ is the size of the feature space. It can be either a \pkg{pandas} data frame or a \pkg{numpy} array.
    \item \code{training\_events}: Vector of event indicators, of size $n$. It corresponds to the vector $(\delta_1, \dots, \delta_n)\tr$ and it must be a \proglang{Python} list, a \pkg{pandas} series or a \pkg{numpy} array. 
    \item \code{training\_times}: Vector of event times, of size $n$. It corresponds to the vector $(\tau_1, \dots, \tau_n )\tr$ and this must be a \proglang{Python} list, a \pkg{pandas} series or a \pkg{numpy} array.
    \item \code{model\_output\_times} (optional): Vector of times for which the model provides a prediction, as explained is Section \ref{sec:survival_models}. By default, the vector of distinct times $(t_1, \dots, t_{m+1})\tr$, obtained from \code{training\_times}, is used. If provided, it must be a \pkg{numpy} array.
    \item \code{H0} (optional): Vector of baseline cumulative hazard values, of size $m+1$, used by the local Cox Proportional Hazards Model. If the user provides it, then a \pkg{numpy} array, a \proglang{Python} list or a \code{StepFunction} (from \pkg{sksurv} package) must be used. If the user does not provide it, the package uses the non-parametric algorithm of Nelson-Aalen \citep{NelsonAalen}. It is computed using the events $\delta_i$ (\code{training\_events}) and times $\tau_i$ (\code{training\_times}).
    \item \code{kernel\_width} (optional): Bandwidth of the kernel ($b$ parameter defined in Section \ref{sec:neighbour_kernel}) used in the neighbours generating process as well as to compute the vector of weights $\mathbf{w}$. A float must be used. The default value for this parameter is equal to $4/(n[p+2])]^{1/(p+4)}$. See Section \ref{sec:neighbour_kernel} for more details.
    \item \code{functional\_norm} (optional): Norm used in order to calculate the distance between the logarithm of the Cox Proportional Hazards Model, $\phi_{cox}(t)$, and the logarithm of the black box model, $\phi(t)$. If provided, it must be either a float $k \geq 1$, in order to use $\ell^k$,  or the string ``inf'', in order to use $\ell^{\infty}$. The default value is set to 2. See Section \ref{sec:functional_norm} for more details.
    \item \code{random\_state} (optional): Number to be used for the random seed. The user must provide a value if the results obtained must be reproducible every time the code is executed. The default is set to empty (no reproducibility needed).
\end{itemize}

In order to obtain the coefficients of the local Cox Proportional Hazards Model, the aforementioned class has a specific method:
\begin{CodeInput}
    explain_instance(
        data_row, predict_fn,
        type_fn, num_samples, verbose
    )
\end{CodeInput}
\begin{itemize}
    \item \code{data\_row}: Instance to be explained, i.e., $\mathbf{x}_*$. It must be a \proglang{Python} list, a \pkg{numpy} array or a \pkg{pandas} series. The length of this array must be equal to the number of columns of the \code{training\_features} matrix, i.e., $p$.
    \item \code{predict\_fn}: Prediction function, i.e., $\hat{H}  \colon \mathbb{R}^p \to \mathbb{R}_{>0}^{q+1}$. It must be a callable (i.e., a \proglang{Python} function). See Section \ref{sec:survival_models} for more details.
    \item \code{type\_fn} (optional): String indicating whether the prediction function, \code{predict\_fn},  is for the Cumulative Hazard Function or for the Survival Function. The default value is set to ``cumulative''. The other option is ``survival''.
    \item \code{num\_samples} (optional): Number of neighbours $N$ to be generated. The default value is set to 1000. See Section \ref{sec:neighbour_kernel} for more details. 
    \item \code{verbose} (optional): Boolean indicating whether to show the \pkg{cvxpy} messages. Default is set to false. 
\end{itemize}

In addition to the main functions, there are three additional functionalities provided by the package. The first one, \fct{plot\_weights}, allows to visualise the SurvLIME coefficients. This function returns a bar plot of the computed values. The function has two optional input parameters. The first one, \code{with\_colour}, is a boolean parameter indicating whether to use a red colour palette for the features that increase the Cumulative Hazard Function and a blue palette for those that decrease it. If it is set to false, the grey colour is used for all the bars. The default value is true. The other input parameter is \code{figure\_path}. In case the user provides a value, it must be a path where the plot is stored as a .png file.

The second functionality is devoted to perform a Monte-Carlo simulation. When using the \fct{explain\_instance} method, the optimisation problem is solved once: a single set of neighbours is generated and, therefore, a single vector of coefficients is obtained. For a given individual $\mathbf{x}_*$, the method \fct{montecarlo\_explanation} allows to obtain a set of vectors (of coefficients) $\{\boldsymbol{\hat{\beta}}_1, \dots, \boldsymbol{\hat{\beta}}_b\}$  each corresponding to a different random set of neighbours. 
In order to use it, the number of simulations, $b$, must be provided. Once all the simulations are performed, the mean value, $\boldsymbol{\bar{\beta}} = 1/b \sum_{j=1}^b\boldsymbol{\hat{\beta}}_j$, is calculated to obtain a single vector of feature importance for the individual $\mathbf{x}_*$.

This method allows to use a matrix $\mathbf{X}_*$ (of size $h \times p$, where $h$ is the number of individuals to be explained) as input, instead of a single individual $\mathbf{x}_*$. Therefore, a matrix $\mathbf{B}$ (of size $h \times p$) is obtained: a row $i$ of $\mathbf{B}$ is a vector containing the feature importance of the individual $i$ of $\mathbf{X}_*$. The function \fct{montecarlo\_explanation} is part of the \class{SurvLimeExplainer} class.
\begin{CodeInput}
    montecarlo_explanation(
        data, predict_fn, type_fn,
        num_samples, num_repetitions, verbose
    )
\end{CodeInput}
Note that all the input parameters are the same as the input parameters of \fct{explain\_instance} except for two of them:
\begin{itemize}
    \item \code{data}: Instances to be explained, i.e., $\mathbf{X}_*$. It must be a \pkg{pandas} data frame, a \pkg{pandas} series,  a \pkg{numpy} array or a \proglang{Python} list.
    \item \code{num\_repetitions} (optional): Integer indication the number of simulations, $b$. The default value is set to 10.
\end{itemize}

Finally, \fct{plot\_montecarlo\_weights} is the last functionality we have developed and it allows to create a boxen plot from the values obtained by \fct{montecarlo\_explanation} method. \fct{plot\_montecarlo\_weights} has two optional input parameters: \code{with\_colour} and \code{figure\_path}. These parameters behave in the same way as the input parameters of the function \fct{plot\_weights}.

\subsection{Code example}
\label{sec:example}
The following code fragment shows how to use the package to compute the importance vector for the features for a single individual. In order to run it, let us suppose we have a machine learning model already trained, denoted by \code{model}, which has a method that obtains a prediction for the Cumulative Hazard Function,  \code{model.predict\_cumulative\_hazard\_function} and it has an attribute containing the times for which the previous method provides a prediction, \code{model.event\_times\_} (we are adopting the notation of \pkg{sksurv} package).

The individual to be explained is denoted by \code{individual}, the dataset containing the features is denoted by \code{features}, the vector containing the event indicators is denoted by \code{events} and the vector containing the times is denoted by \code{times}.
\begin{CodeInput}
    from survlimepy import SurvLimeExplainer
    explainer = SurvLimeExplainer(
            training_features=features,
            training_events=events,
            training_times=times,
            model_output_times=model.event_times_
    )
\end{CodeInput}
\begin{CodeInput}
    explanation = explainer.explain_instance(
        data_row=individual,
        predict_fn=model.predict_cumulative_hazard_function,
        num_samples=1000
    )
    explainer.plot_weights()
\end{CodeInput}

\begin{figure}[!ht]
    \centering
    \includegraphics[width=0.55\textwidth]{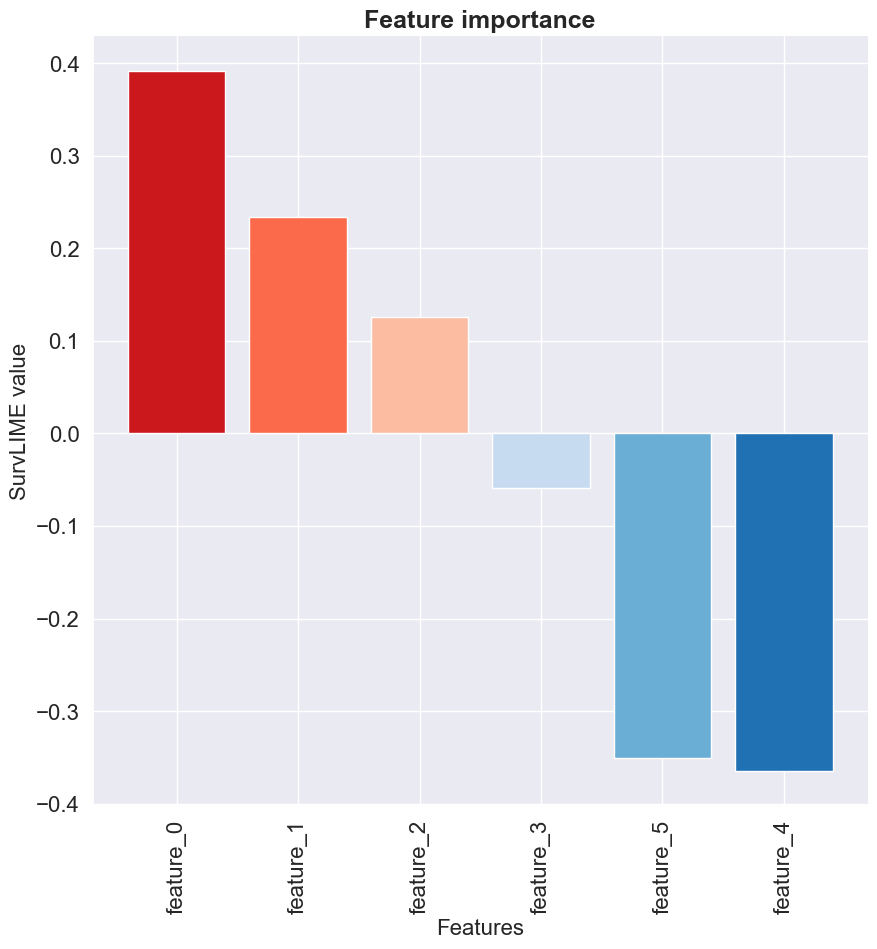}
    \caption{SurvLIME values obtained with \fct{explainer.plot\_weights} method. The input parameter \code{with\_colour} is set to true.}
    \label{fig:code_example}
\end{figure}

The last line displays the importance of each feature. The result is shown in Figure \ref{fig:code_example}. The computed coefficients are displayed in descending order, with a red colour palette for the features that increase the Cumulative Hazard Function and a blue palette for those that decrease it. The remaining input parameters in \class{SurvLimeExplainer} as well as in function \fct{explain\_instance} use their corresponding default values.

The next code block exemplifies how to use \fct{montecarlo\_explanation} to obtain a set of SurvLIME values as well as the \fct{plot\_montecarlo\_weights} method to display them. We make use of the same notation as before, i.e., \code{model.predict\_cumulative\_hazard\_function}, \code{model.event\_times\_}, \code{features}, \code{events} and \code{times}.

\begin{figure}[!ht]
    \centering
    \includegraphics[width=0.55\textwidth]{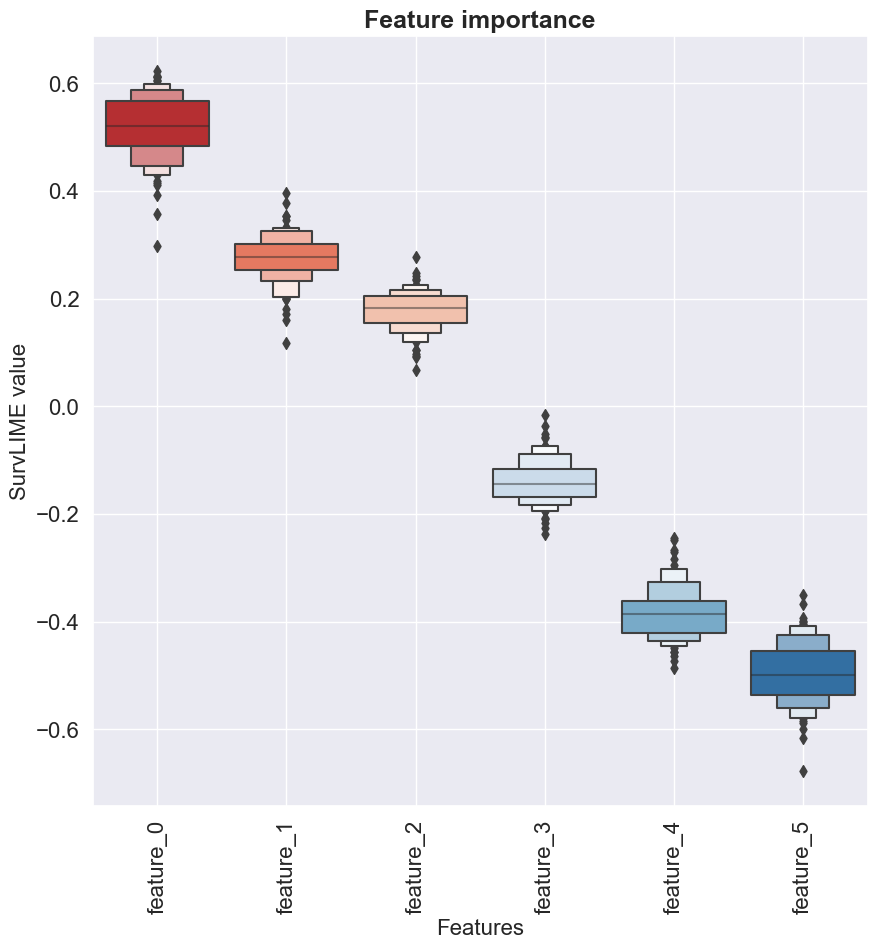}
    \caption{SurvLIME values obtained by means of using the method \fct{explainer.plot\_montecarlo\_weights}. The input parameter \code{with\_colour is set to true.}}
    \label{fig:code_example_mc}
\end{figure}

Instead of explaining a single individual, we explain a set of $h$ individuals. Let, \code{X\_ind} be a \pkg{numpy} array of size $h \times p$. For each individual, we perform 100 repetitions and, for each repetition, 1000 neighbours are generated. The code needed to obtain the results is very similar to the previous one. The last line of the code example is responsible for displaying Figure \ref{fig:code_example_mc}. Note that the variable \code{mc\_explanation} is a \pkg{numpy} array of size $h \times p$, where the row $i$ contains the feature importance for individual $i$ in \code{X\_ind}.
\newpage
\begin{CodeInput}
    from survlimepy import SurvLimeExplainer
    explainer = SurvLimeExplainer(
            training_features=features,
            training_events=events,
            training_times=times,
            model_output_times=model.event_times_
    )
    mc_explanation = explainer.montecarlo_explanation(
        data=X_ind,
        predict_fn=model.predict_cumulative_hazard_function,
        num_repetitions=100,
        num_samples=1000
    )
    explainer.plot_montecarlo_weights()
\end{CodeInput}

\section{Experiments}
\label{sec:experiments}
In this section, we present the experiments performed to test the implementation of our package \pkg{SurvLIMEpy}. In order to ensure reproducibility we have created a separate repository\footnote{\url{https://github.com/imatge-upc/SurvLIME-experiments}} in which we share the code used throughout this section. 

We conduct two types of experiments. The first is by means of simulated data, as the authors of the original paper of SurvLIME. Given that they describe in detail how their data was generated, we are able to follow the same procedure. As we use simulated data we can compare the results of the SurvLIME algorithm with the data generating process. Therefore, we can  measure how much the coefficients provided by the algorithm deviate from the real coefficients (i.e. the simulated ones).

The second set of experiments is with real survival datasets. In this part, we use machine learning as well as deep learning algorithms. Our goal is to show how \pkg{SurvLIMEpy} can be used with the state-of-the-art machine learning models. For those experiments, we do not have results to compare with, unlike what happens in the case of simulated data. Therefore, just qualitative insights are provided.

\subsection{Simulated data}
\label{sec:simulated_data}
First, two sets of data are generated randomly and uniformly in the $p$-dimensional sphere, where $p=5$. Each set is configured as follows:
\begin{itemize}
    \item Set 1: Center, $c_1 = (0, 0, 0, 0, 0)$, radius, $r_1 = 8$, number of individuals, $n_1 = 1000$.
    \item Set 2: Center, $c_2 = (4, -8, 2, 4, 2)$, radius, $r_2 = 8$, number of individuals, $n_2 = 1000$.
\end{itemize}

Using these parameters, two datasets  represented by the matrices $\mathbf{X}_r$ of size $n_r \times p$ are generated ($r \in \{1,2\}$). A row from these datasets represents an individual and a column represents a feature. Therefore, $x_{ij}$ represents the value of feature $j$ for the individual $i$.

The Weibull distribution is used to generate time data \citep{survival_data}. This distribution respects the assumption of proportional hazards, the same as the Cox Proportional Hazards Model does. The Weibull distribution is determined by two parameters: the scale, $\lambda$, and the shape, $\nu$. Given the set of data $r$, a vector of time to events (of size $n_r$) is generated as
\begin{equation}
    \label{eq:time_to_event}
    \boldsymbol\tau_r = \left(\frac{-\ln(\mathbf{u}_r)}{\lambda_r\exp(\mathbf{X}_r\boldsymbol\beta_r)}\right)^{1/\nu_r},
\end{equation}
where $\mathbf{u}_r$ is a vector of $n_r$ independent and identically uniform distributions in the interval $(0,1)$. Both functions, the logarithm and the exponential, are applied component-wise. As done in \cite{survlime}, all times greater than 2000 are constrained to 2000. Each set $r$ has the following set of parameters:
\begin{itemize}
    \item Set 1: $\lambda_1 = 10^{-5}$, $\nu_1 = 2$, $\boldsymbol\beta_1\tr = (10^{-6}, 0.1, -0.15, 10^{-6}, 10^{-6})$.
    \item Set 2: $\lambda_2 = 10^{-5}$, $\nu_2 = 2$, $\boldsymbol\beta_2\tr = (10^{-6}, -0.15, 10^{-6}, 10^{-6}, -0.1)$.
\end{itemize}

Note that for the first set, the second and the third features are the most important ones. On the other hand, for the second set, the second and the fifth features are the most relevant.

In order to generate the event indicator, a Bernoulli distribution, with a probability of success equal to 0.9, is used. For each set a vector (of size $n_r$) of independent and identically distributed random variables is obtained. Let $\boldsymbol\delta_r$ be the vector of such realisations. The random survival data of each set $r$ is represented by a triplet 
$\mathbf{D}_r = (\mathbf{X}_r, \boldsymbol\tau_r, \boldsymbol\delta_r)$.

Even though the authors of the original SurvLIME paper simulated data this way, it is worth mentioning that this is not the standard procedure in Survival Analysis. 
The usual way to generate data consists of using two different distributions of times, $\boldsymbol\tau_0$ and $\boldsymbol\tau_1$: $\boldsymbol\tau_0$ is the censoring time  
and $\boldsymbol\tau_1$ is the time-to-event. 
Then, the vector of observed times $\boldsymbol\tau=(\tau_i)$ is obtained as  $\tau_i = \min(\tau_{0i}, \tau_{1i})$. In order to generate the event indicator vector $\boldsymbol{\delta}=(\delta_i)$, it is taken into account both vectors $\boldsymbol\tau_0$ and $\boldsymbol\tau_1$: $\delta_i=\mathds{1}_{\{\tau_{1i} \leq \tau_{0i}\}}$. In this way, it is obtained that $\Prob(\delta_i = 1) = \Prob(\tau_{1i} \leq \tau_{0i})$. Nonetheless, we proceed in the same way as in the original paper so that the results can be compared.

\pkg{SurvLIMEpy} allows to create a random survival dataset according to the criteria described previously. The class 
\class{RandomSurvivalData} manages this part.
\newpage
\begin{CodeInput}
   RandomSurvivalData(
        center, radius, coefficients, prob_event,
        lambda_weibull, v_weibull, time_cap, random_seed
    )
\end{CodeInput}

\iffalse
\begin{CodeInput}
   RandomSurvivalData(center, radius, coefficients, prob_event, lambda_weibull, 
   v_weibull, time_cap, random_seed)
\end{CodeInput}
\fi

\begin{itemize}
    \item \code{center}: The center of the set. It must be a \proglang{Python} list of length $p$.
    \item \code{radius}: The radius of the set. It must be a float.
    \item \code{coefficients}: The $\boldsymbol\beta_r$ vector that is involved in Expression (\ref{eq:time_to_event}). It must be a \proglang{Python} list of length $p$.
    \item \code{prob\_event}: The probability for the Bernoulli distribution. It must be a float in $(0, 1)$.
    \item \code{lambda\_weibull}: The $\lambda_r$ parameter that is involved in Expression (\ref{eq:time_to_event}). It must be a float positive number.
    \item \code{v\_weibull}: The $\nu_r$ parameter that is involved in Expression (\ref{eq:time_to_event}). It must be a float positive number.
    \item \code{time\_cap} (optional): If the time obtained is greater than \code{time\_cap}, then \code{time\_cap} is used. It must be a float positive number.
    \item \code{random\_seed} (optional): Number to be used for the random seed. The user must provide a value if the results obtained must be reproducible every time the code is executed. The default is set to empty (no reproducibility needed).
\end{itemize}

This class contains the method \code{random\_survival\_data(num\_points)} that returns the dataset. The input parameter, \code{num\_points}, is an integer indicating the number of individuals, $n_r$, to generate. The output of this function is a tuple of three objects: (1) $\mathbf{X}_r$ the matrix containing the features (of size $n_r \times p$); (2) $\boldsymbol\tau_r$ the vector of times to event (of size $n_r$); (3) $\boldsymbol\delta_r$ the vector of event indicators (of size $n_r$).

After obtaining both datasets, they are split randomly into two parts, a training dataset, $\mathbf{D}_r^{train}$ and a test dataset, $\mathbf{D}_r^{test}$. The training dataset consists of 900 individuals, whereas the test dataset consists of 100 individuals.

For each training dataset, a Cox Proportional Hazards Model is fitted. Let $\hat{H}_r(\mathbf{x}, t)$, $r \in \{1, 2\}$, be the resulting models. The next step is to use \pkg{SurvLIMEpy} to obtain the importance of each feature. The test datasets, still unexploited, are used to rank the relevance of each feature. For a given test individual from set $r$, the set up for \pkg{SurvLIMEpy} is:

\begin{itemize}
    \item Training dataset,
    $\mathbf{D} = \mathbf{D}_r^{train}$.
    \item Number of neighbours, $N_r=1000$.
    \item Black-box model for the Cumulative Hazard Function: $\hat{H}_r(\mathbf{x}, t)$.
    \item Kernel function, 
    $\Ker(\cdot, \cdot)=$ Gaussian Radial Basis function.
\end{itemize}

\begin{figure*}[!htbp]
    \centering
    \begin{minipage}{0.49\textwidth}
        \includegraphics[width=1\textwidth]{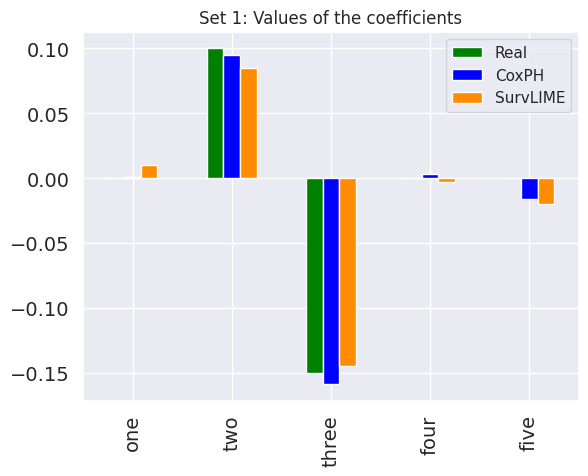}
    \end{minipage}
    \begin{minipage}{0.49\textwidth}
        \includegraphics[width=1\textwidth]{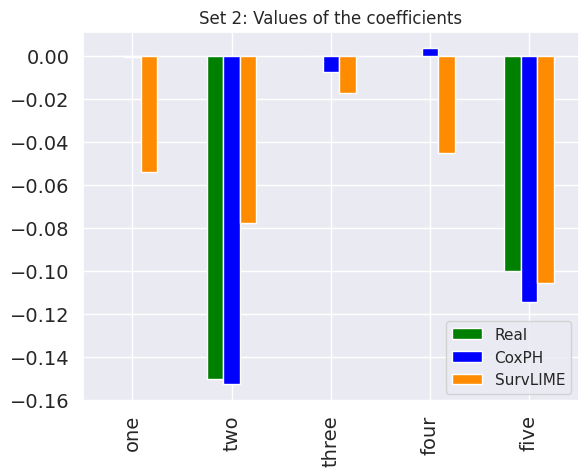}
    \end{minipage}
    \caption{
        Real coefficients for parameters (green), estimated coefficients by CoxPH (blue) and SurvLIME coefficients (orange). Results for set 1 (left). Results for set 2 (right). The individual to be explained is the center of the set.
    }
    \label{fig:ground_truths_exp1}
\end{figure*}

Figure \ref{fig:ground_truths_exp1} shows the results obtained using \pkg{SurvLIMEpy} package to compute the coefficients. In green, the vector of real coefficients, $\boldsymbol\beta_r$, is depicted. In blue, the estimated parameters according to Cox Proportional Hazards Model,
$\hat{\boldsymbol\beta}_r^{c}$. In orange, the coefficients obtained by \pkg{SurvLIMEpy},
$\hat{\boldsymbol\beta}_r^{s}$, $r \in \{1, 2\}$. The individual to be explained is the center of the set. Note that the results we have obtained are similar to the ones obtained in the original paper of SurvLIME.

Given that the real coefficients, $\boldsymbol\beta_r$, are known, the $\ell^2$ distance between $\boldsymbol\beta_r$ and $\hat{\boldsymbol\beta}_r^{s}$ can be computed. In order to study the variance of SurvLIME algorithm, the previous experiment is repeated 100 times, i.e, a Monte-Carlo simulation is performed. Throughout all the simulations, the individual to be explained is the same, the center of the set.

Thus, a set of 100 distances are obtained, $\{d_1, \dots, d_{100}\}$. From this set, the mean, the minimum and the maximum distance can be calculated. Let
$\hat{\boldsymbol\beta}_{mean}^{s}$, 
$\hat{\boldsymbol\beta}_{min}^{s}$ and
$\hat{\boldsymbol\beta}_{max}^{s}$
be the SurvLIME coefficients related to those distances. Doing such a Monte-Carlo simulation for all the individuals in the test datasets, $\mathbf{D}_1^{test}$ and $\mathbf{D}_2^{test}$, leads to obtain 3 different samples of SurvLIME coefficients: 
$\{\hat{\boldsymbol\beta}_{mean,1}^{s}, \dots, \hat{\boldsymbol\beta}_{mean,100}^{s}\}$,
$\{\hat{\boldsymbol\beta}_{min,1}^{s}, \dots, \hat{\boldsymbol\beta}_{min,100}^{s}\}$ and
$\{\hat{\boldsymbol\beta}_{max,1}^{s}, \dots, \hat{\boldsymbol\beta}_{\max,100}^{s}\}$.

\begin{figure*}[!ht]
    \begin{minipage}{\textwidth}
    \begin{minipage}{0.32\textwidth}
        \includegraphics[width=1\textwidth]{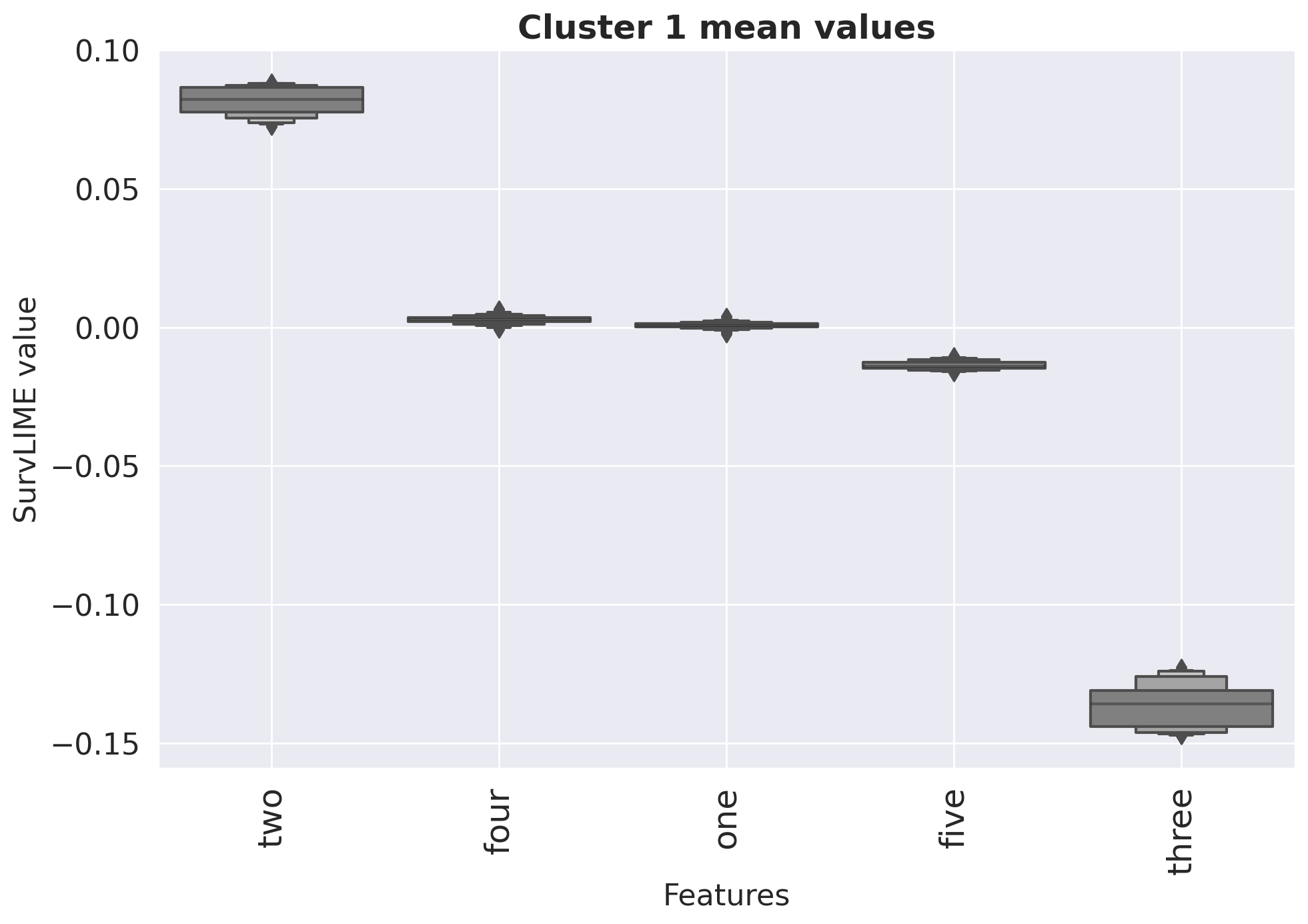}
    \end{minipage}
    \begin{minipage}{0.32\textwidth}
        \includegraphics[width=1\textwidth]{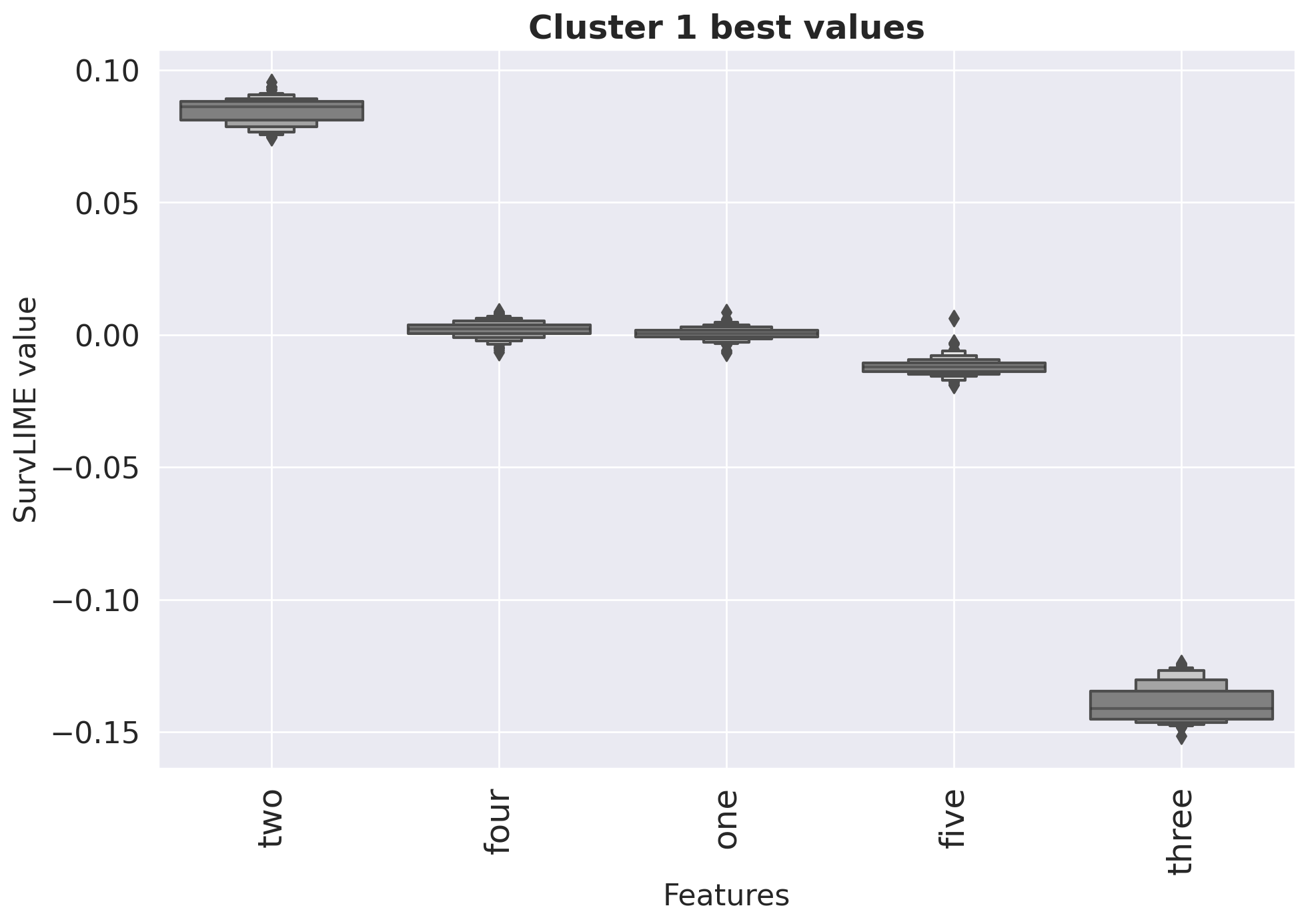}
    \end{minipage}
    \begin{minipage}{0.32\textwidth}
        \includegraphics[width=1\textwidth]{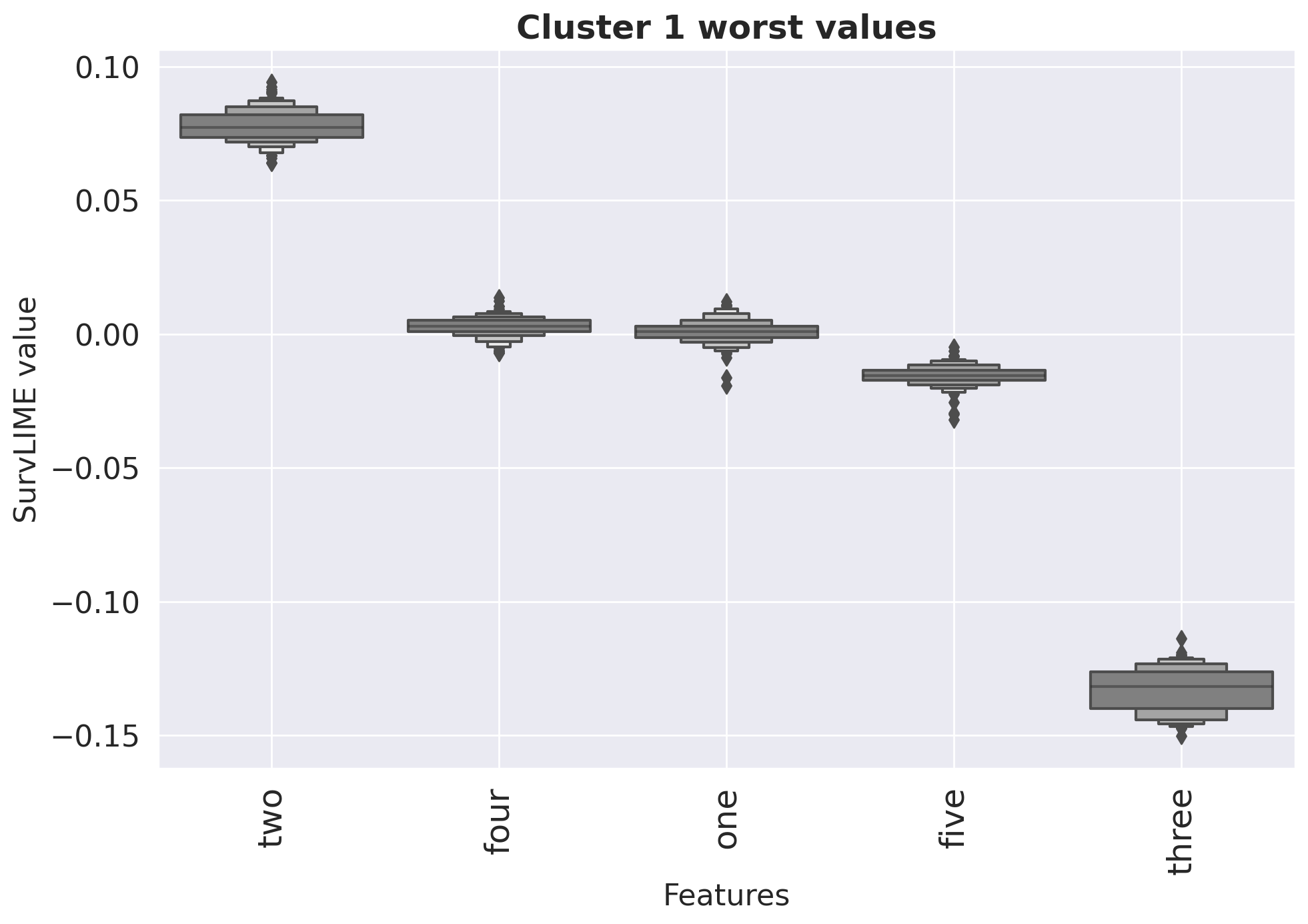}
    \end{minipage}
    \end{minipage}
    \vspace{3mm}
    
    \begin{minipage}{\textwidth}
    \begin{minipage}{0.32\textwidth}
        \includegraphics[width=1\textwidth]{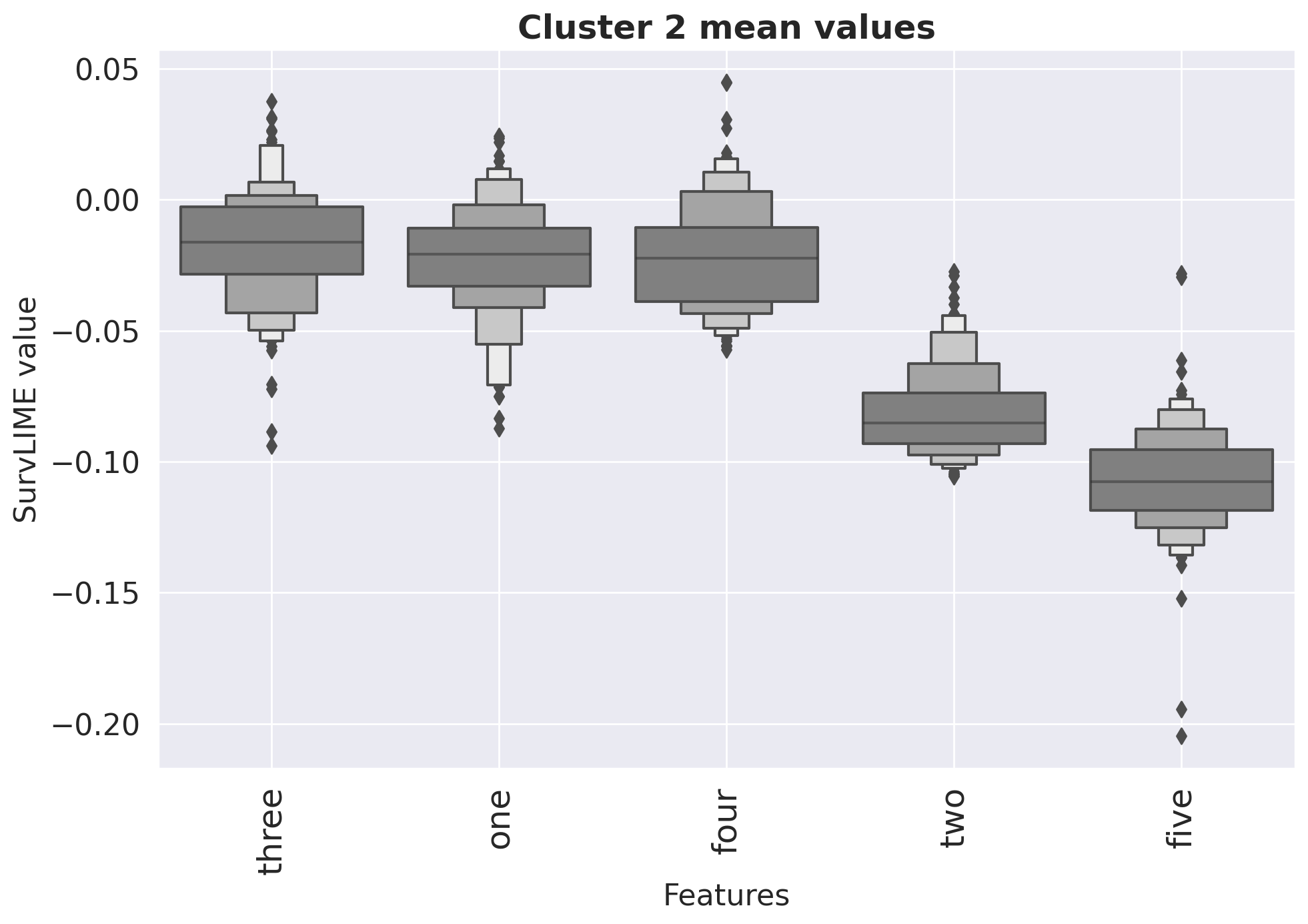}
    \end{minipage}
    \begin{minipage}{0.32\textwidth}
        \includegraphics[width=1\textwidth]{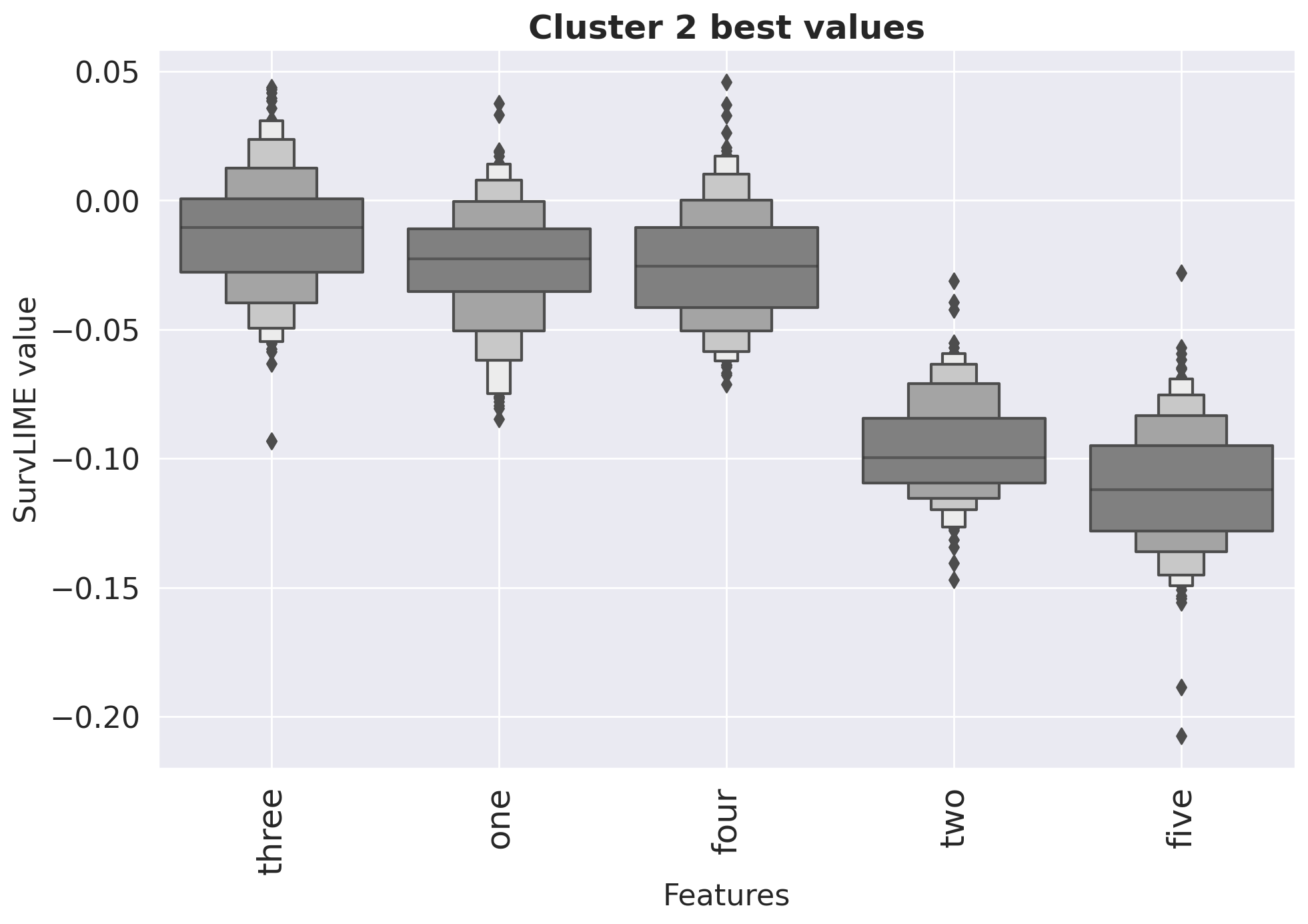}
    \end{minipage}
    \begin{minipage}{0.32\textwidth}
        \includegraphics[width=1\textwidth]{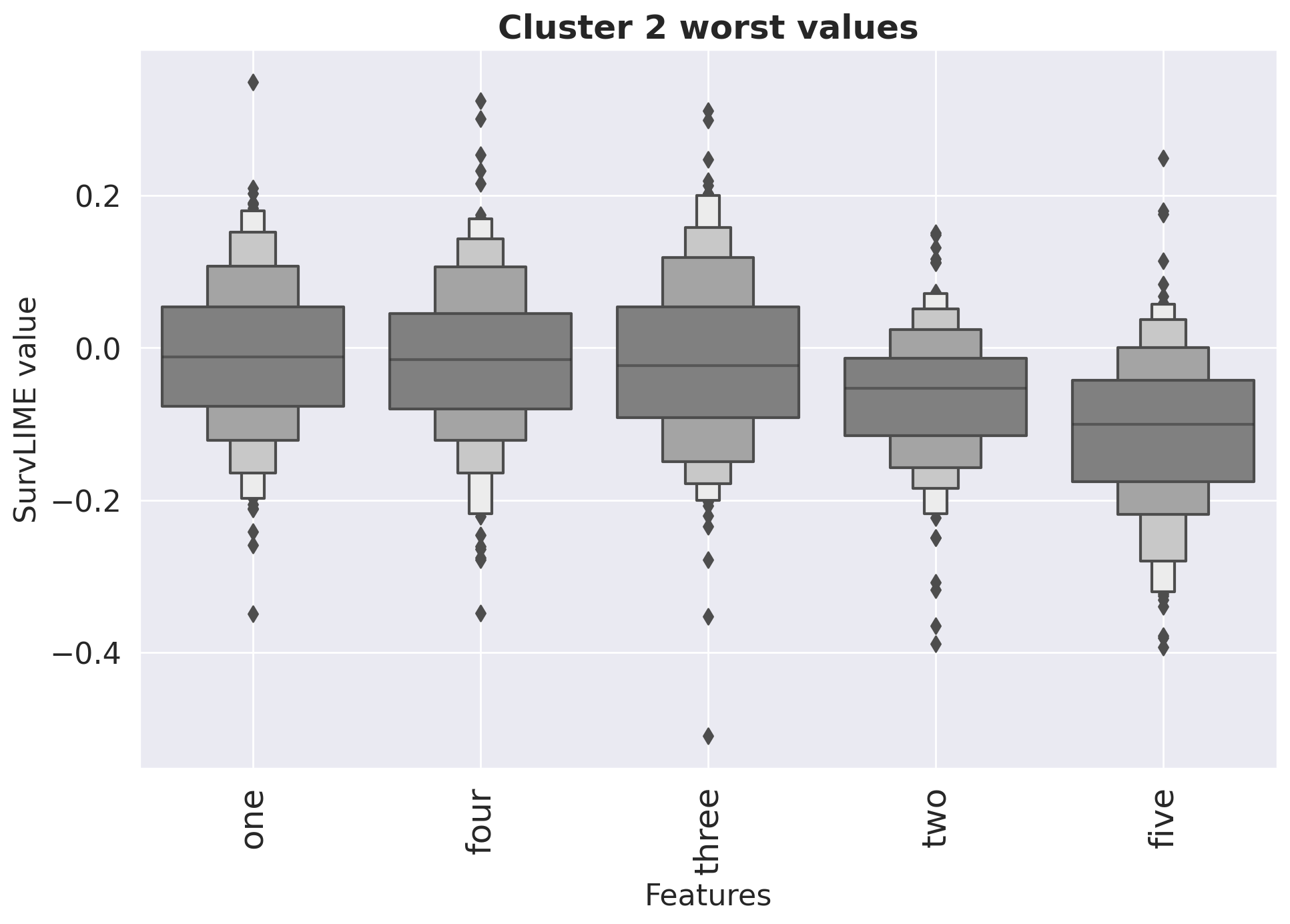}
    \end{minipage}
    \end{minipage}
    \caption{
        Boxen plot for the mean (left) minimum (middle) and maximum (right) distance. Results are shown for individuals of the first set (top) and the second set (bottom).
    }
   \label{fig:exp_1.1}
\end{figure*}

Figure \ref{fig:exp_1.1} shows the boxen plots for the three previous sets of coefficients. The left plots depict the boxen plot for the mean coefficient; the middle plots are for the minimum coefficient; the right ones correspond to the maximum coefficient. The results show that the coefficients of SurvLIME were close to the real coefficients for both sets of data. Furthermore, the mean values of the computed coefficients behave similarly to the best approximations and they show a low  variance. 

In the worst case scenario, SurvLIME does not behave as well as in the other two scenarios. The variance of the SurvLIME coefficients is much higher,
especially for the second set of data. However, the bias is as good as the bias of the other two scenarios.

\subsection{Real data}
\label{sec:real_data}
Now, we test our implementation on three open-access datasets. 
Each dataset is presented together with a bivariate analysis. 
For categorical features, the percentage of individuals that experienced the event is computed for each category.
Continuous features are categorised according to their quartiles, and the resulting categorical features are described as before.

The first dataset is the UDCA dataset \citep{udca_article}. It contains individuals with primary biliary cirrhosis (PBC) that were randomised for treatment with ursodeoxycholic acid (UDCA). A total of $9.46\%$ of the individuals experienced the event. The features of this dataset are:
\begin{itemize}
    \item trt (categorical): Treatment received. 0 is for placebo and 1 is for UDCA.
    \item stage (categorical): Stage of disease. 0 is for better and 1 is for worse.
    \item bili (continuous): Bilirubin value at entry.
    \item riskscore (continuous): The Mayo PBC risk score at entry.
\end{itemize}
Note that the UDCA dataset contains an individual whose riskscore is missing. We drop this individual from the dataset. 
The bivariate descriptive analysis is displayed in Table \ref{tab:udca_analysis}.
\begin{table}[!ht]
    \begin{center}
        \begin{tabular}{@{}l r@{}}
            \toprule
            \multicolumn{2}{c}{trt feature} \\ 
            Category & percentage\_cat \\ [0.5ex]
            \midrule
            0 & 11.90\\
            1 & 7.06\\
            \bottomrule
        \end{tabular}
        \hfil   %<---
        \begin{tabular}{@{}l r@{}}
            \toprule
            \multicolumn{2}{c}{stage feature} \\
            
            Category & percentage\_cat \\ [0.5ex]
            \midrule
            0 & 3.85\\
            1 & 12.00\\
            \bottomrule
        \end{tabular}
    
        \medskip
        \begin{tabular}{@{}l r@{}}
            \toprule
            \multicolumn{2}{c}{bili feature} \\
            Category & percentage\_cat \\ [0.5ex]
            \midrule
            $\leq 0.6$ & 2.17\\
            $(0.6, 1]$ & 2.56\\
            $(1, 1.9]$ & 14.30\\
            $>1.9$ & 19.00\\
            \bottomrule
        \end{tabular}
        \hfil   %<---
        \begin{tabular}{@{}l r@{}}
            \toprule
            \multicolumn{2}{c}{riskscore feature} \\
            
            Category & percentage\_cat \\ [0.5ex]
            \midrule
            $\leq 4.3$ & 0.00\\
            $(4.3, 5]$ & 0.00\\
            $(5, 5.8]$ & 9.52\\
            $>5.8$ & 30.80\\
            \bottomrule
        \end{tabular}
    \end{center}
    \caption{Percentage of individuals that have experienced the event according to each category for all the features in the UDCA dataset.}
    \label{tab:udca_analysis}
\end{table}
 
The second dataset is the LUNG dataset \citep{Lung_dataset}. It contains individuals with advanced lung cancer from the North Central Cancer Treatment Group. A total of $70.33\%$ of the individuals experienced the event. The features of this dataset are:
\begin{itemize}
    \item inst (categorical): Institution code. The institutions are coded with numbers between 1 and 33.
    \item sex (categorical): Gender. 1 is for male and 2 is for female.
    \item ph.ecog (categorical): ECOG performance score as rated by the physician. The categories are:
    \begin{itemize}
        \item 0: Asymptomatic.
        \item 1: Symptomatic but completely ambulatory.
        \item 2: In bed <50\% of the day.
        \item 3: In bed > 50\% of the day but not bedbound.
    \end{itemize}
    \item age (continuous): Age of the individual.
    \item ph.karno (continuous): Karnofsky performance score rated by physician.
    \item pat.karno (continuous): Karnofsky performance score as rated by the individual.
    \item meal.cal (continuous): Calories consumed at meals.
    \item wt.loss (continuous): Weight loss in last six months.
\end{itemize}

We drop some information regarding LUNG dataset. First, we do not use the feature inst because it does not provide any further information allowing institutions identification. 
Second, we remove the meal.cal feature, since it contains a total of 20.6\% of missing values. Third, 18 individuals have at least one feature with missing information. We drop those individuals from the dataset. Finally, with regards the feature ph.ecog, just a single individual is in the category 3. We do not consider this individual, therefore we drop it. After this preprocessing, we are left with 209 individuals.

As for the UDCA dataset, a bivariate descriptive analysis is performed in LUNG dataset. Table \ref{tab:lung_analysis} contains the results. Those features dropped from the dataset are not included in that table.

\begin{table}[!ht]
 \begin{minipage}{0.99\textwidth}
        \centering
        \begin{tabular}{@{}p{2cm} r@{}}
            \toprule
            \multicolumn{2}{c}{sex feature} \\ 
            Category & percentage\_cat \\ [0.5ex]
            \midrule
            1 & 79.80\\
            2 & 56.50\\
            \bottomrule
        \end{tabular}
        \hfil   %<---
        \begin{tabular}{@{}p{2cm} r@{}}
            \toprule
            \multicolumn{2}{c}{ph.ecog feature} \\
            Category & percentage\_cat \\ [0.5ex]
            \midrule
            0 & 56.70\\
            1 & 71.70\\
            2 & 86.00\\
            \bottomrule
        \end{tabular}
    \vspace{3mm}
    \end{minipage}
    \begin{minipage}{0.99\textwidth}
     \centering
        \begin{tabular}{@{}p{2cm} r@{}}
            \toprule
            \multicolumn{2}{c}{age feature} \\
            Category & percentage\_cat \\ [0.5ex]
            \midrule
            $\leq 56$ & 64.80\\
            $(56, 63]$ & 65.40\\
            $(63, 69]$ & 70.60\\
            $>69$ & 80.80\\
            \bottomrule
        \end{tabular}
        \hfil   %<---
        \begin{tabular}{@{}p{2cm} r@{}}
            \toprule
            \multicolumn{2}{c}{ph.karno feature} \\
            Category & percentage\_cat \\ [0.5ex]
            \midrule
            $\leq 80$ & 77.00\\
            $(80, 90]$ & 62.70\\
            $>90$ & 62.10\\
            \bottomrule
        \end{tabular}
        \medskip
   \end{minipage}
   
   \begin{minipage}{0.99\textwidth}
             \centering
        \begin{tabular}{@{}p{2cm} r@{}}
            \toprule
            \multicolumn{2}{c}{pat.karno feature} \\
            Category & percentage\_cat \\ [0.5ex]
            \midrule
            $\leq 70$ & 80.00\\
            $(70, 80]$ & 75.00\\
            $(80, 90]$ & 62.70\\
            $>90$ & 56.20\\
            \bottomrule
        \end{tabular}
        \hfil   %<---
        \begin{tabular}{@{}p{2cm} r@{}}
            \toprule
            \multicolumn{2}{c}{wt.loss feature} \\
            Category & percentage\_cat \\ [0.5ex]
            \midrule
            $\leq 0$ & 68.90\\
            $(0, 6]$ & 59.10\\
            $(6, 15]$ & 79.20\\
            $>15$ & 72.50\\
            \bottomrule
        \end{tabular}
\end{minipage}
    \caption{Percentage of individuals that have experienced the event according to each category for all the features in the LUNG dataset.}
    \label{tab:lung_analysis}
\end{table}

The last dataset is the Veteran dataset \citep{Veterans_dataset} which consists of individuals with advanced inoperable Lung cancer. The individuals were part of a randomised trial of two treatment regimens. The event of interest for the three datasets is the individual's death. A total of $93.43\%$ of the individuals experienced the event. The features of this dataset are:
\begin{itemize}
    \item trt (categorical): Treatment received. 1 is for standard and 2 is for test.
    \item prior (categorical): It indicates if the patient has received another therapy before the current one. 0 means no and 10 means yes.
    \item  celltype(categorical): Histological type of the tumor. The categories are: squamous, smallcell, adeno and large.
    \item karno (continuous): Karnofsky performance score.
    \item age (continuous): Age of the individual.
    \item diagtime (continuous): Months from diagnosis to randomisation.
\end{itemize}
Note that the Veteran dataset does not contain any missing value. The results of the bivariate descriptive analysis for the Veteran dataset are displayed in Table \ref{tab:veteran_analysis}.
\begin{table}[!ht]
    \begin{center}
        \begin{tabular}{@{}p{2cm} r@{}}
            \toprule
            \multicolumn{2}{c}{trt feature} \\ 
            Category & percentage\_cat \\ [0.5ex]
            \midrule
            1 & 92.80\\
            2 & 94.10\\
            \bottomrule
        \end{tabular}
        \hfil   %<---
        \begin{tabular}{@{}p{2cm} r@{}}
            \toprule
            \multicolumn{2}{c}{prior feature} \\
            Category & percentage\_cat \\ [0.5ex]
            \midrule
            0 & 93.80\\
            10 & 92.50\\
            \bottomrule
        \end{tabular}
    
        \medskip
        \begin{tabular}{@{}p{2cm} r@{}}
            \toprule
            \multicolumn{2}{c}{celltype feature} \\
            Category & percentage\_cat \\ [0.5ex]
            \midrule
            squamous & 88.60\\
            smallcell & 93.80\\
            adeno & 96.30\\
            large & 96.30\\
            \bottomrule
        \end{tabular}
        \hfil   %<---
        \begin{tabular}{@{}p{2cm} r@{}}
            \toprule
            \multicolumn{2}{c}{karno feature} \\
            Category & percentage\_cat \\ [0.5ex]
            \midrule
            $\leq 40$ & 97.40\\
            $(40, 60]$ & 95.10\\
            $(60, 75]$ & 92.00\\
            $>75$ & 87.90\\
            \bottomrule
        \end{tabular}

        \medskip
        \begin{tabular}{@{}p{2cm} r@{}}
            \toprule
            \multicolumn{2}{c}{age feature} \\
            Category & percentage\_cat \\ [0.5ex]
            \midrule
            $\leq 51$ & 94.30\\
            $(51, 62]$ & 87.20\\
            $(62, 66]$ & 100.00\\
            $>66$ & 93.30\\
            \bottomrule
        \end{tabular}
        \hfil   %<---
        \begin{tabular}{@{}p{2cm} r@{}}
            \toprule
            \multicolumn{2}{c}{diagtime feature} \\
            Category & percentage\_cat \\ [0.5ex]
            \midrule
            $\leq 3$ & 90.50\\
            $(3, 5]$ & 97.00\\
            $(5, 11]$ & 90.00\\
            $>11$ & 96.9\\
            \bottomrule
        \end{tabular}
    \end{center}
    \caption{Percentage of individuals that have experienced the event according to each category for all the features in the Veteran dataset.}
    \label{tab:veteran_analysis}
\end{table}

Table \ref{tab:open_access_datasets} shows a brief summary of each dataset: $p$ corresponds to the number of features, while $p^{*}$ is the number of  features after pre-processing (dropping and doing one-hot-encoding), $n$ denotes the number of individuals of the dataset, and $n_{full}$ is the number of individuals once the missing values are dropped.

\begin{table}[!ht]
    \centering
    \begin{tabular}{@{}p{0.52\linewidth} p{0.15\linewidth} p{0.05\linewidth} p{0.05\linewidth} p{0.05\linewidth} p{0.05\linewidth}@{}} 
        \toprule
        Dataset & Acronym & $p$ & $p^{*}$ & $n$ & $n_{full}$\\
        \midrule
        Trial of Usrodeoxycholic Acid & UDCA & 4 & 4 & 170 &169 \\
        NCCTG Lung Cancer & LUNG & 8 & 7 & 228 & 209\\
        Veterans' Administration Lung Cancer Study & Veteran & 6 & 8 & 137 & 137\\
        \bottomrule
    \end{tabular}
    \caption{Summary of the open access datasets used, where $p$ is number of features for the corresponding dataset, $p^{*}$ is the number of  features after pre-processing (dropping and doing one-hot-encoding), $n$ denotes the total number of individuals in the dataset, and $n_{full}$ is the number of individuals after dropping missing values.}
    \label{tab:open_access_datasets}
\end{table}

We model the event of interest by means of machine learning algorithms. Given a dataset $\mathbf{D}$, it is divided randomly into two sets: a training dataset, $\mathbf{D}^{train}$, and a test dataset, $\mathbf{D}^{test}$, using $90\%$ of individuals for training and $10\%$ for testing.

Once the data is split, we  preprocess $\mathbf{D}^{train}$. We apply one-hot-encoding to categorical features. If a categorical feature has $k$ categories, then we create $k-1$ binary features. The category without a binary feature is the reference category. After that, the original feature is deleted from the dataset since we use the $k-1$ new features treated as continuous ones. Continuous features are also preprocessed. Given $\mathbf{\tilde{x}}_j$, we first estimate the mean, $\hat{\mu}^j_{train}$, and the standard deviation, $\hat{\sigma}^j_{train}$. Then, the standarisation performed is $(\mathbf{\tilde{x}}_j - \hat{\mu}^j_{train})/\hat{\sigma}^j_{train}$. This new feature is used instead of $\mathbf{\tilde{x}}_j$.

The same preprocess is applied on $\mathbf{D}^{test}$. Note that the parameters that involve the preprocess (for both, categorical and continuous features) are taken from the preprocess performed on $\mathbf{D}^{train}$, i.e., nothing is estimated in the test set. Let $\mathbf{\tilde{D}}^{train}$ and $\mathbf{\tilde{D}}^{test}$ be the datasets obtained after preprocessing them.

Afterwards, a model is trained in $\mathbf{\tilde{D}}^{train}$ and $\mathbf{\tilde{D}}^{test}$ is used  to obtain the c-index value, a goodness-of-fit measure for survival models (see Appendix \ref{appendix:surv-analysis} for more details about c-index and Survival Analysis).

In this section, we use five distinct machine learning algorithms: the Cox Proportional Hazards Model (CoxPH), Random Survival Forest (RSF) (both from \pkg{sksurv} package), eXtreme Gradient Boosted Survival Trees (XGB) (from \pkg{xgbse} package) as well as continuous and time-discrete deep learning models, DeepSurv and DeepHit (both from \pkg{pycox} package). We have performed an hyperparameter tuning strategy for each model and dataset.

Having trained a model, \pkg{SurvLIMEpy} is applied to obtain feature importance. For a given individual $i$ of $\mathbf{\tilde{D}}^{test}$, SurvLIME algorithm is used 100 times, which produces a set of 100 coefficients: $\{\hat{\boldsymbol\beta}_{i,1}^{s}, \dots, \hat{\boldsymbol\beta}_{i,100}^{s}\}$. Then, the mean value across all the simulation is calculated, $\bar{\boldsymbol\beta}_{i}^{s}=(1/100)\sum_{j=1}^{100}\hat{\boldsymbol\beta}_{i,j}^{s}$. That vector, $\bar{\boldsymbol\beta}_{i}^{s}$, is used as the feature importance for the individual $i$. This process is applied to all the individuals in the test dataset. Therefore, a set of coefficients $\{\bar{\boldsymbol\beta}_{1}^{s}, \dots, \bar{\boldsymbol\beta}_{n_t}^{s}\}$ is obtained, where $n_t$ is the total number of individuals in the test dataset. This set of coefficients is used in this study. Note that for UDCA $n_t$ is equal to 17, for LUNG it is equal to 21, and for Veteran it is equal to 14.

\begin{figure*}[!htbp]
\centering
    \begin{minipage}{0.49\textwidth}
        \includegraphics[width=1\textwidth]{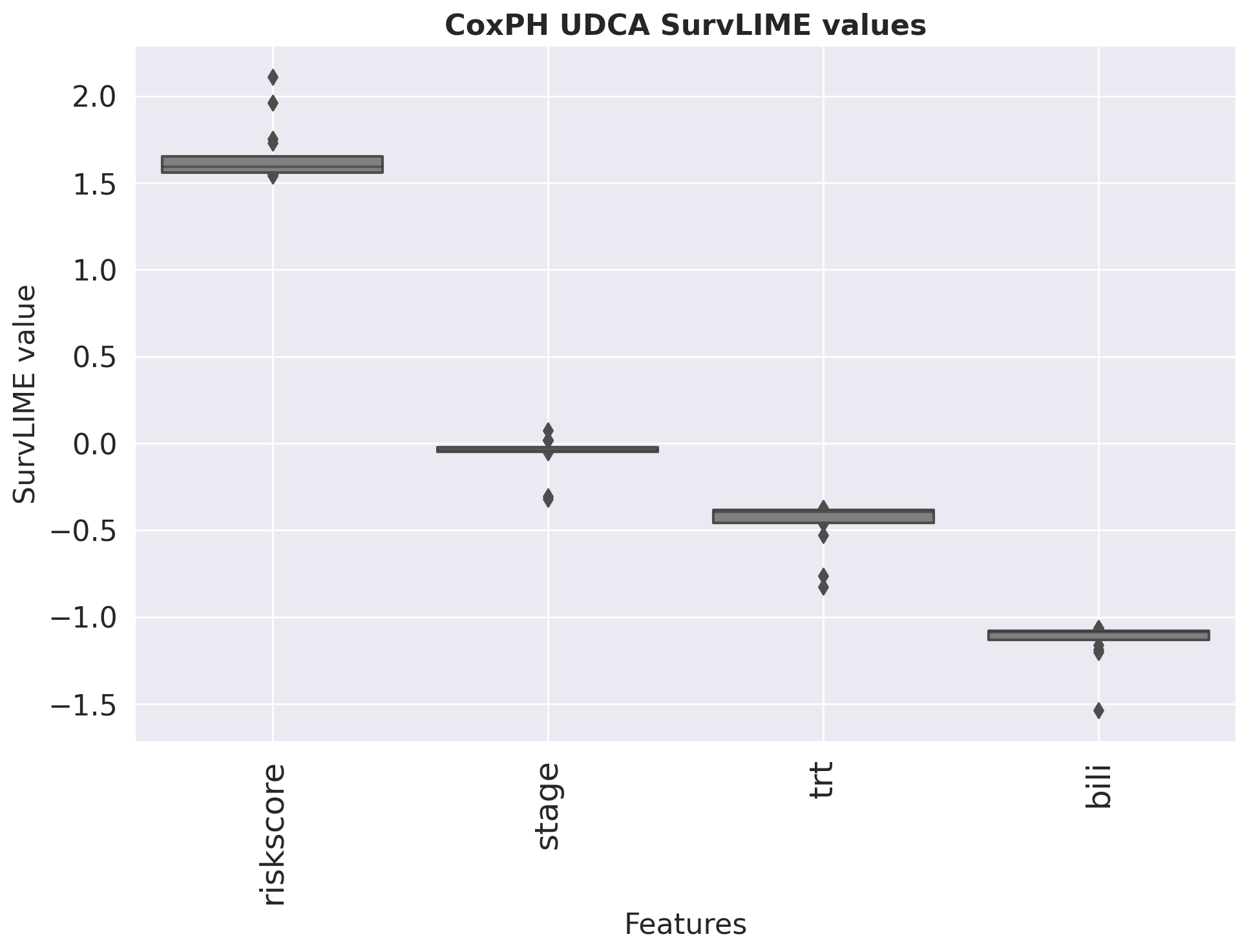}
    \end{minipage}
    \\
    \begin{minipage}{0.49\textwidth}
        \includegraphics[width=1\textwidth]{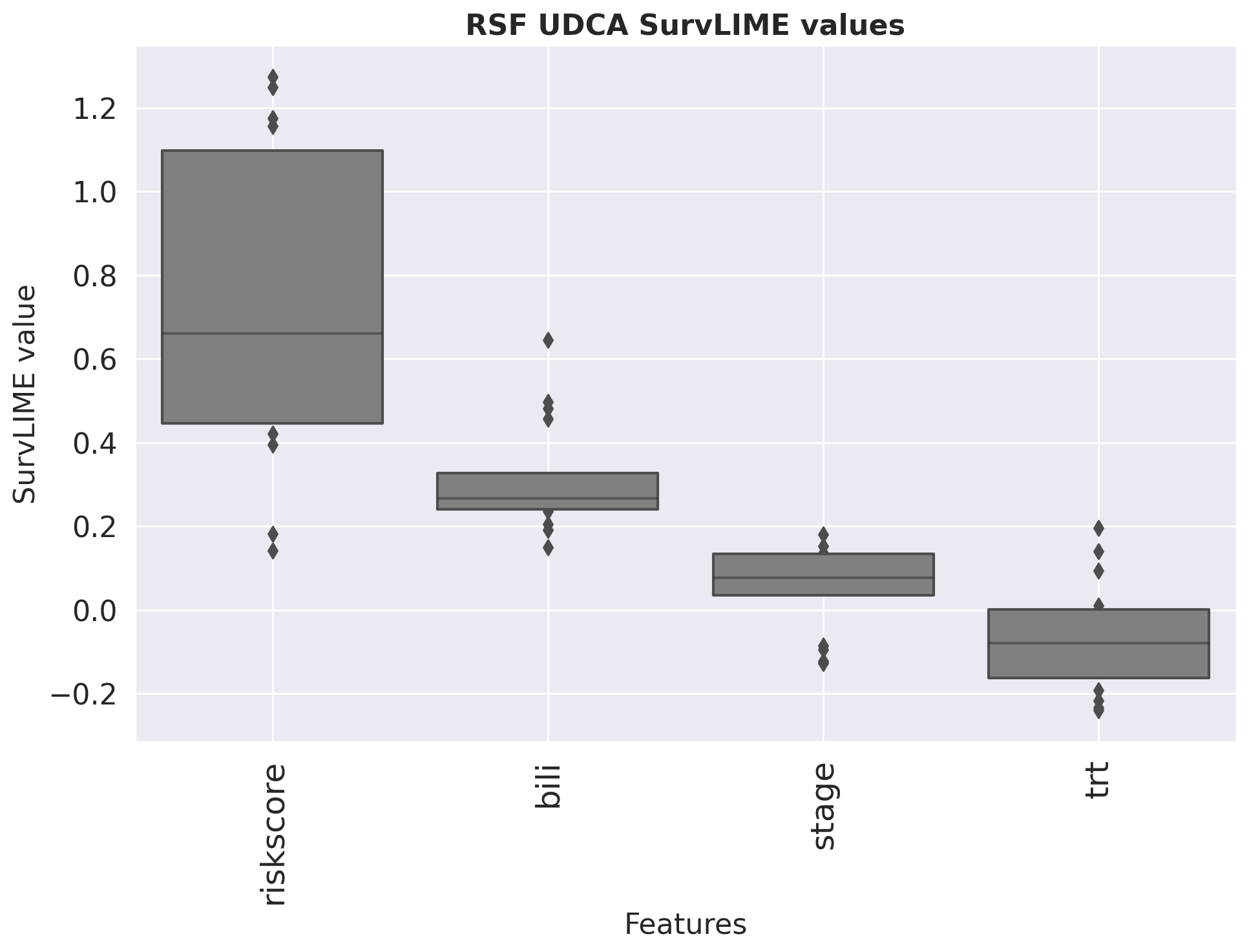}
    \end{minipage}
    \begin{minipage}{0.49\textwidth}
        \includegraphics[width=1\textwidth]{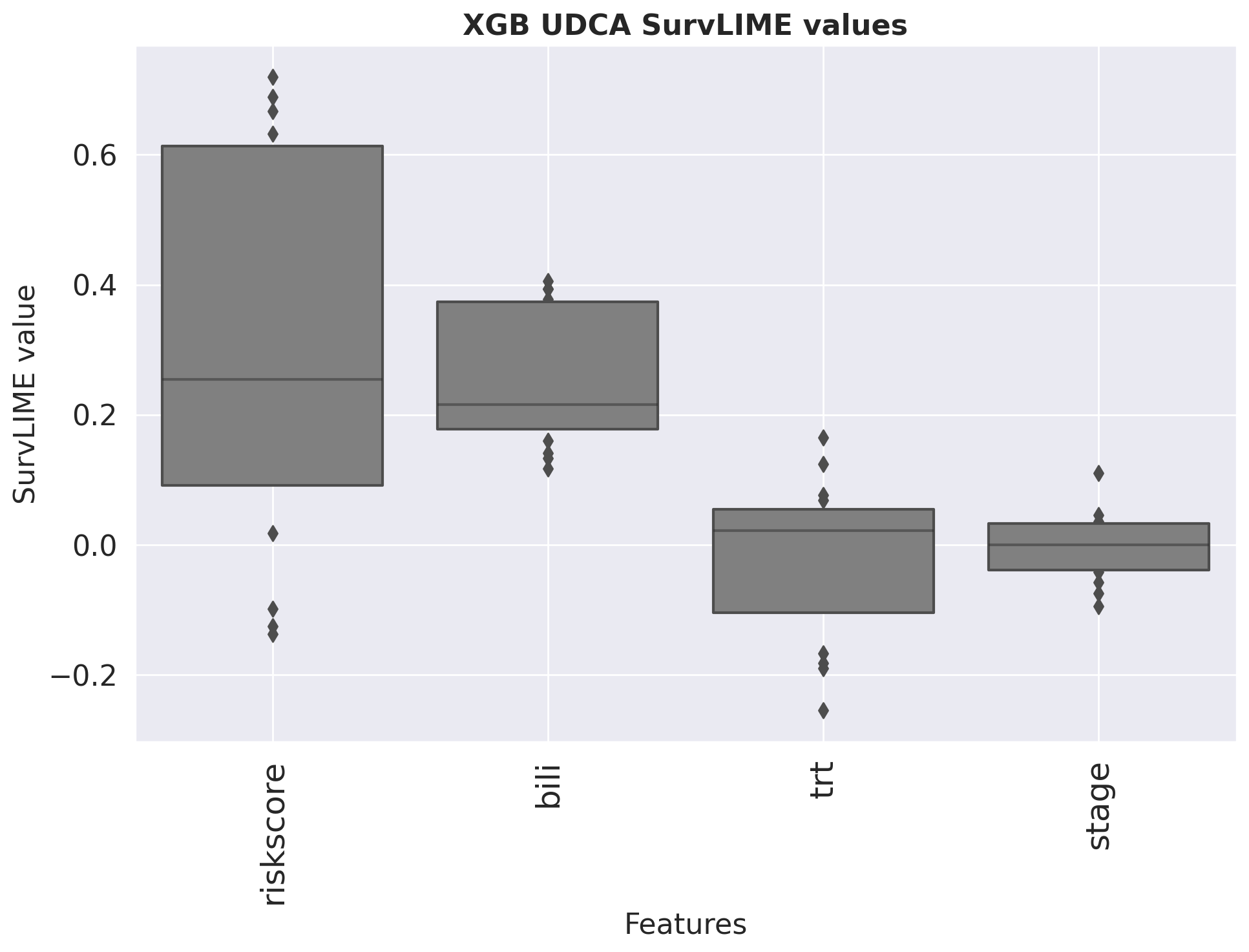}
    \end{minipage}
   \caption{Feature importance for the UDCA dataset. The input parameter \code{with\_colour} is set to false.}
   \label{fig:Udca_weights}
\end{figure*}

\begin{figure*}[!htbp]
 \centering
    \begin{minipage}{0.49\textwidth}
        \includegraphics[width=1\textwidth]{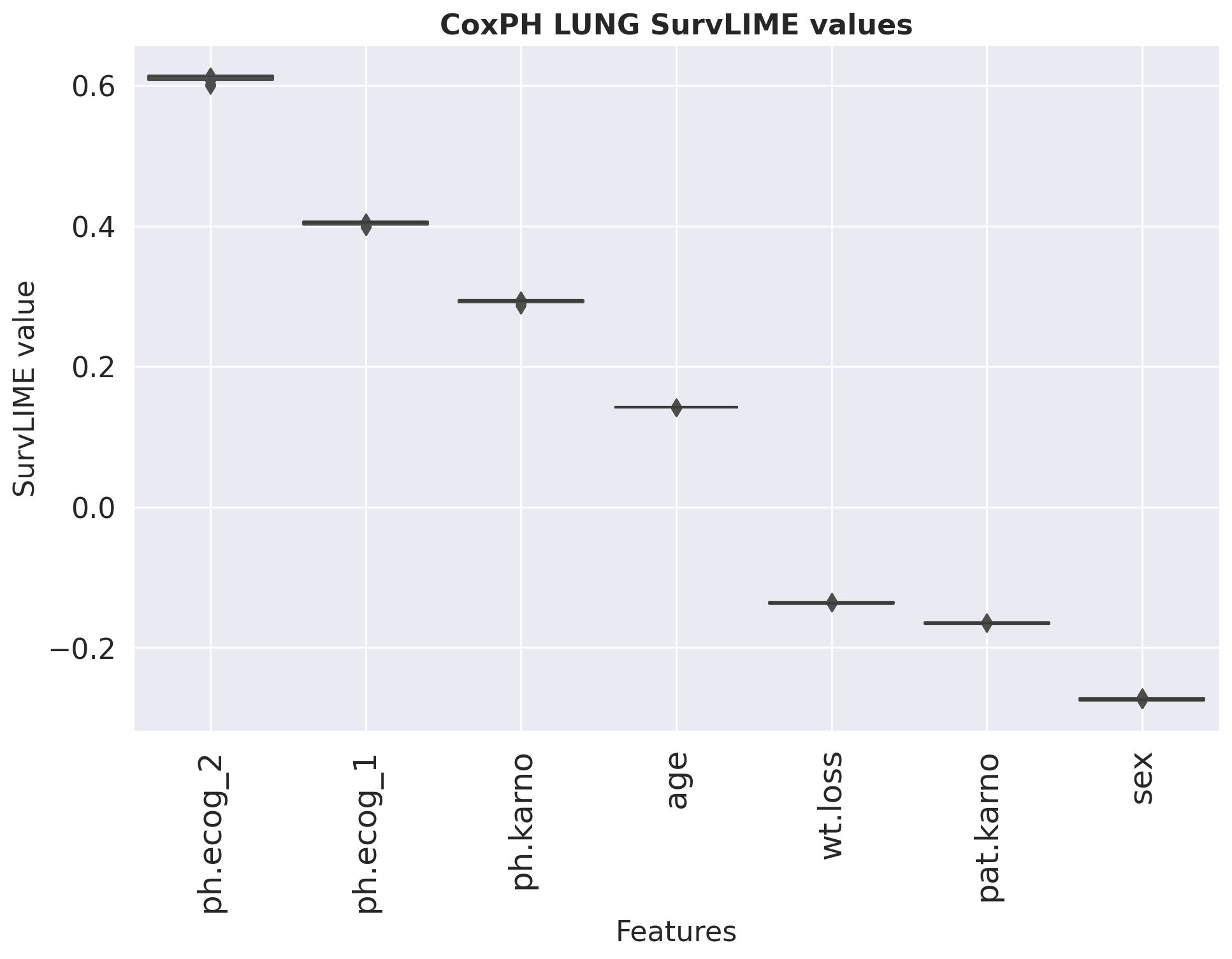}
    \end{minipage}\\
    
    \begin{minipage}{0.49\textwidth}
        \includegraphics[width=1\textwidth]{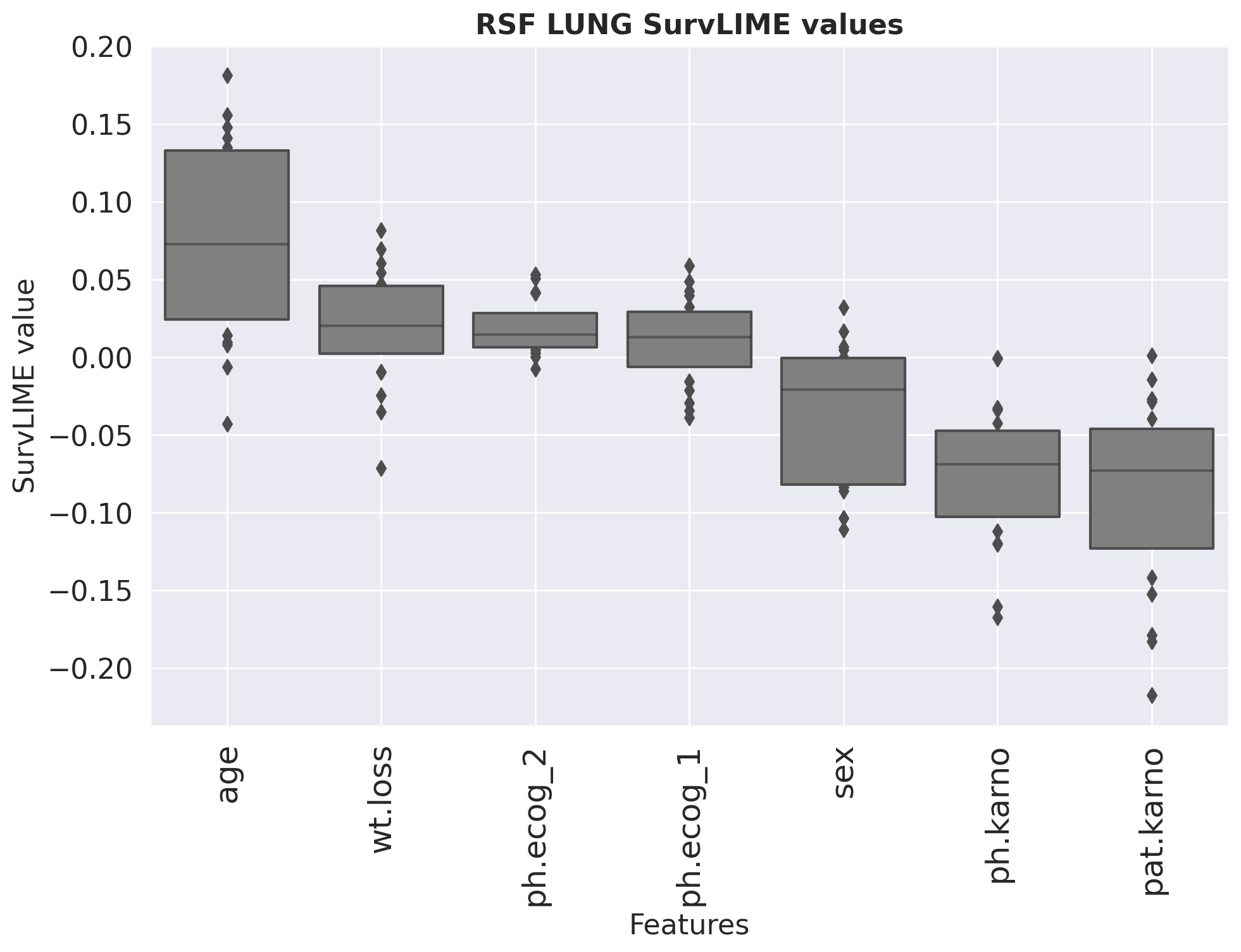}
    \end{minipage}
    \begin{minipage}{0.49\textwidth}
        \includegraphics[width=1\textwidth]{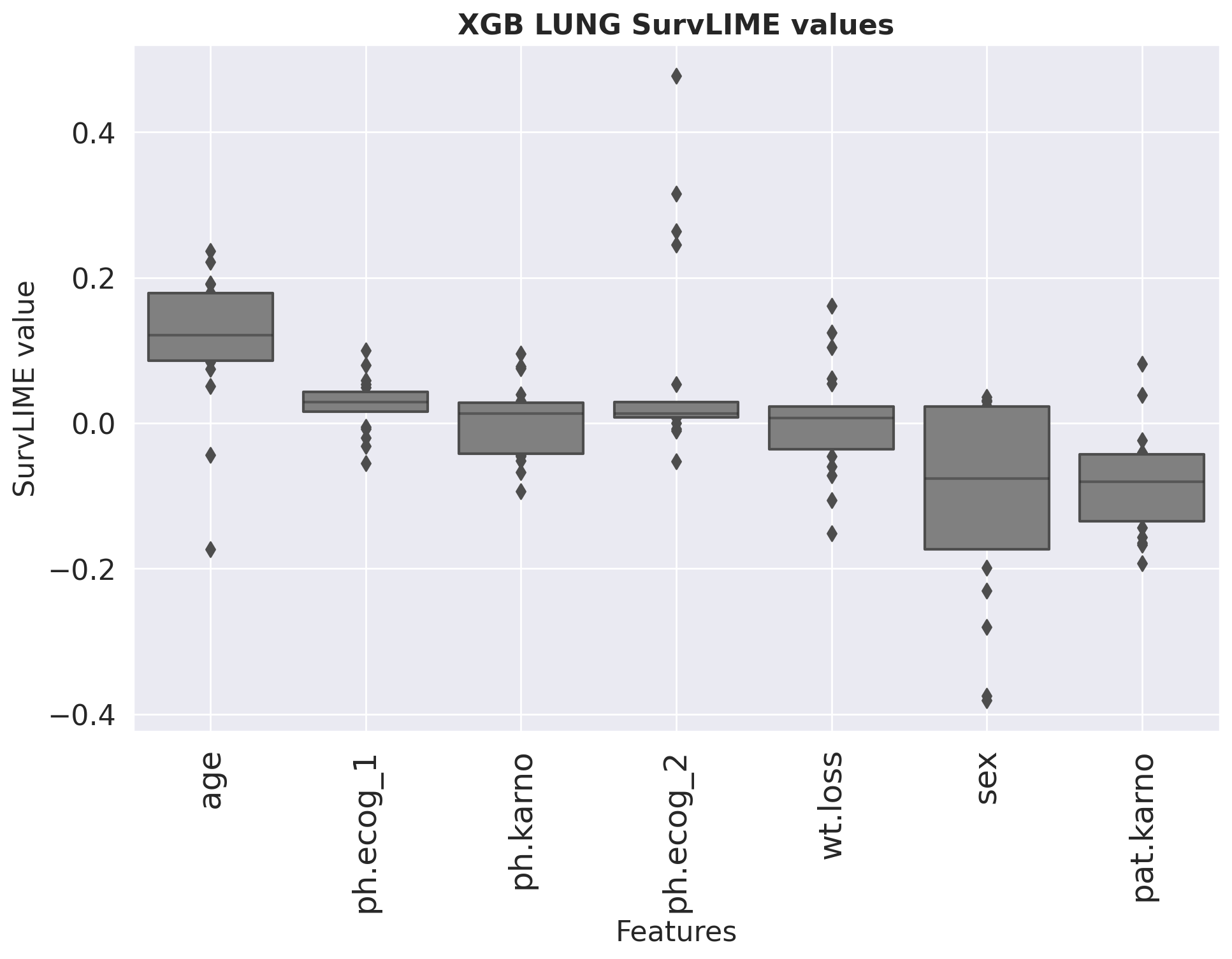}
    \end{minipage}
   \caption{Feature importance for the LUNG dataset. The input parameter \code{with\_colour} is set to false.}
   \label{fig:Lung_weights}
\end{figure*}

\begin{figure*}[!htbp]
    \centering
    \begin{minipage}{0.49\textwidth}
        \includegraphics[width=1\textwidth]{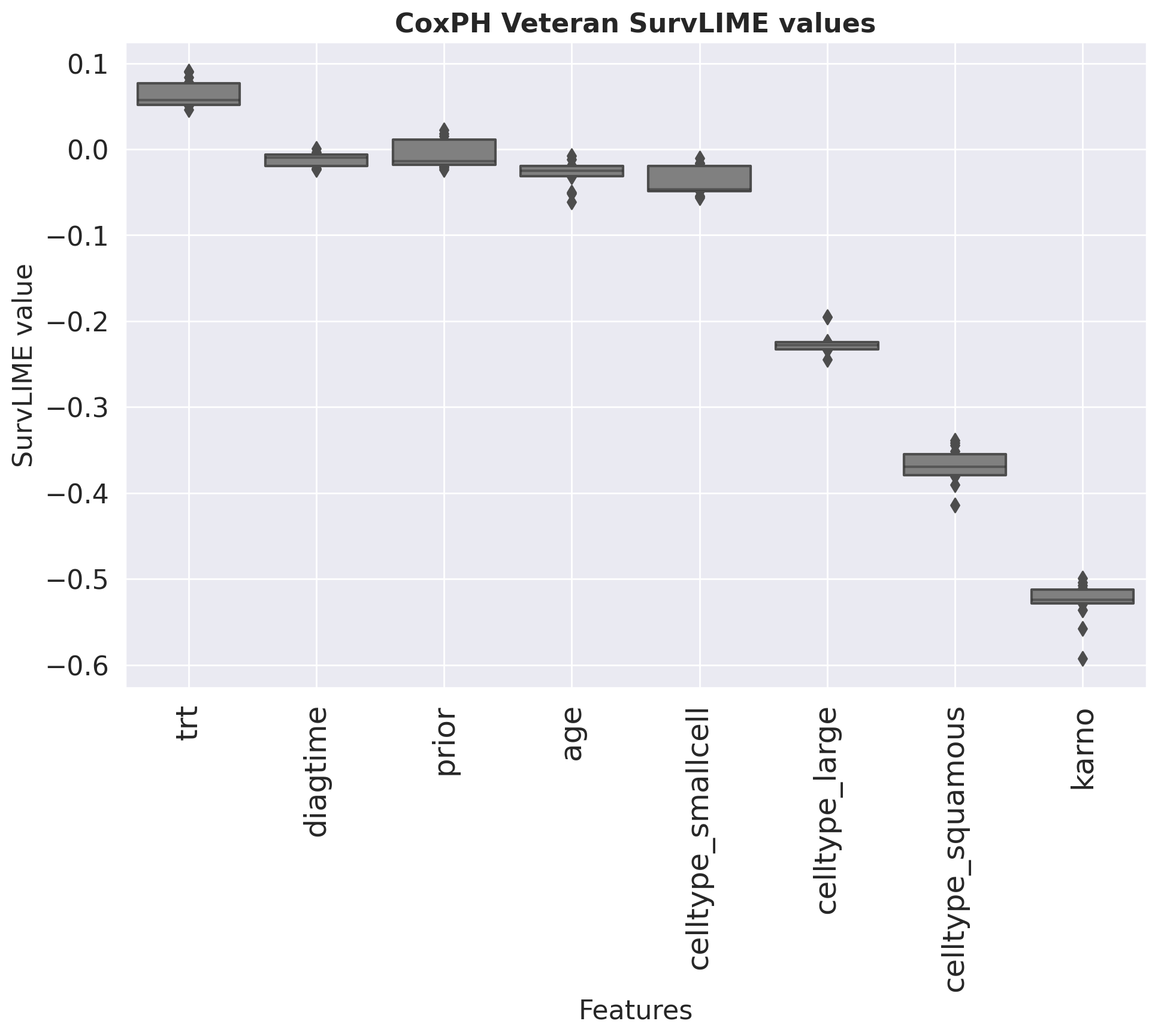}
    \end{minipage}
    \\
    \begin{minipage}{0.49\textwidth}
        \includegraphics[width=1\textwidth]{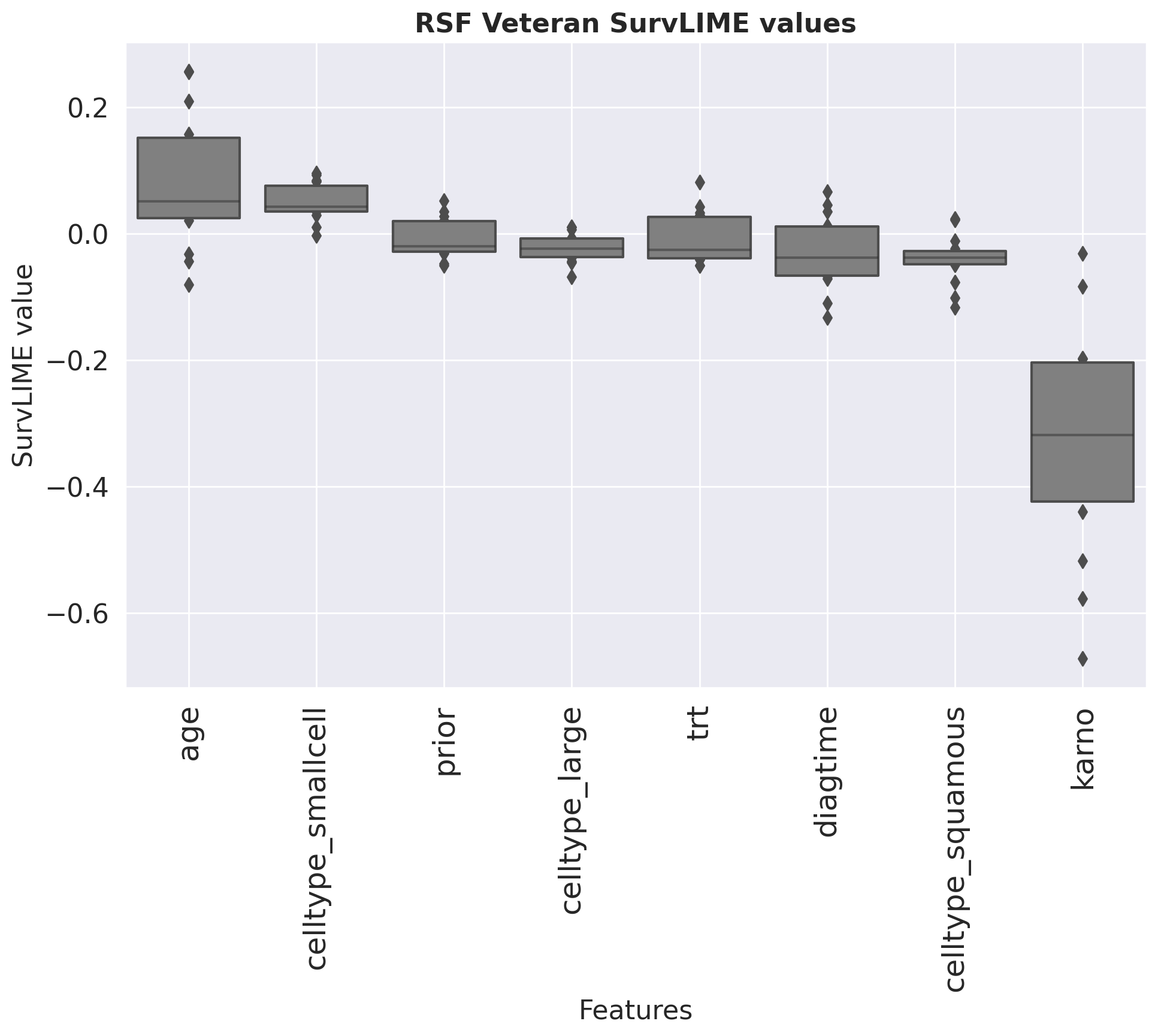}
    \end{minipage}
    \begin{minipage}{0.49\textwidth}
        \includegraphics[width=1\textwidth]{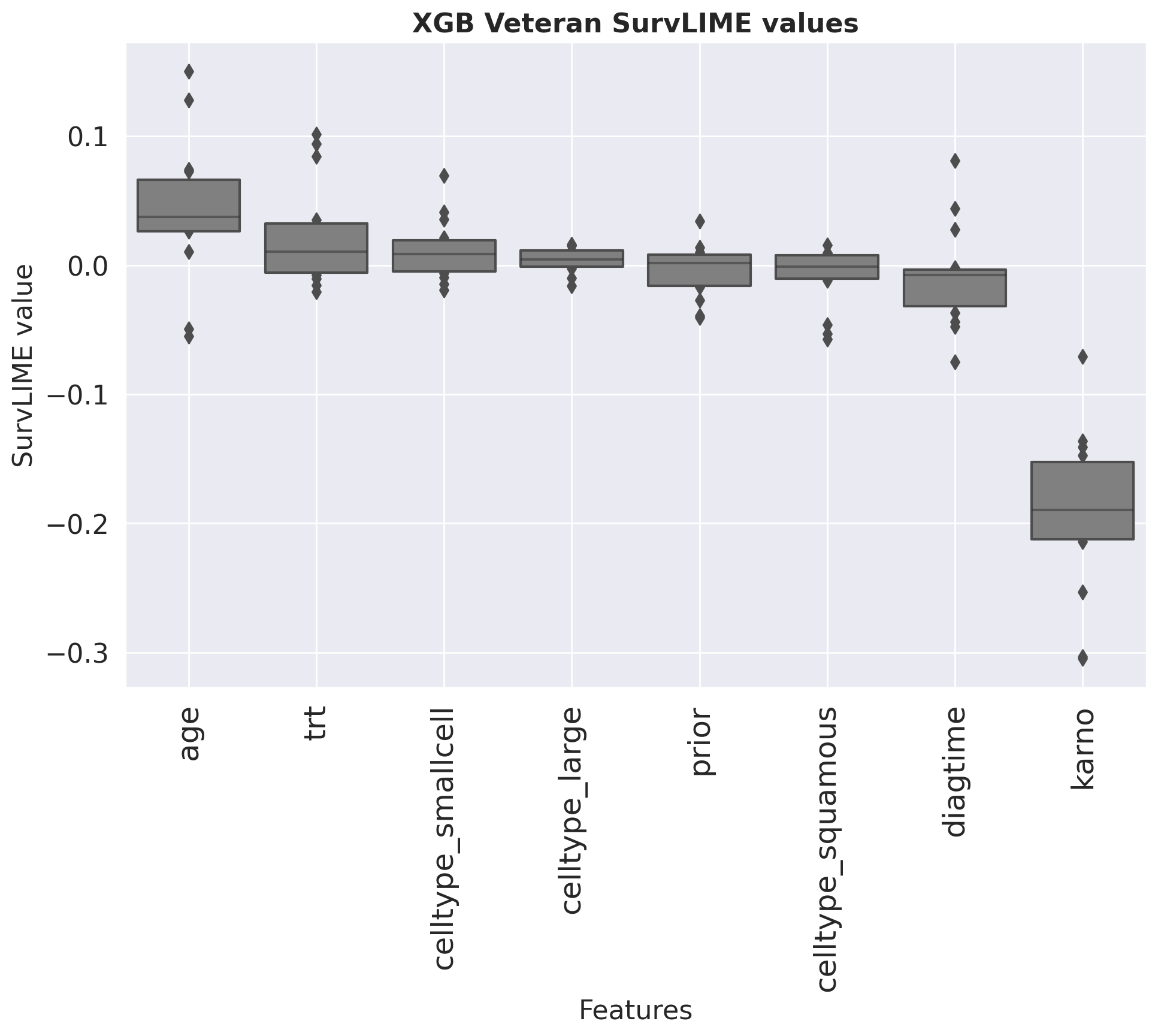}
    \end{minipage}
   \caption{Feature importance for the Veteran dataset. The input parameter \code{with\_colour} is set to false.}
   \label{fig:Veterans_weights}
\end{figure*}

Table \ref{tab:c_index} shows the value of the c-index for the different models. It can be seen that for all the datasets, the c-index related to deep learning models (i.e., DeepSurv and DeepHit) is 0.5 or close to this value, which is the value that one would obtain if a random model were taking decisions. An explanation for such a value is found in the number of individuals: the sample size of the datasets is small relative to the number of parameters of those models. \Cref{fig:Lung_weights,fig:Udca_weights,fig:Veterans_weights} depict the feature importance for each model and dataset. The number of points used to obtain each of the boxen plots depicted in these figures is equal to the number of individuals in $\mathbf{\tilde{D}}^{test}$. For each figure, the set of SurvLIME coefficients used to produce those figures is $\{\bar{\boldsymbol\beta}_{1}^{s}, \dots, \bar{\boldsymbol\beta}_{n_t}^{s}\}$.

As the value of the c-index is so low for DeepSurv and DeepHit, we do not show the feature importance for those models in this section. However, in Section \ref{sec:simulated_data_dl} we use simulated data in order to train deep learning models with an acceptable c-index and show the feature importance for those models.

\begin{table}[!h]
    \centering
    \begin{tabular}{@{}l r r r@{}}
        \toprule
         Model & UDCA & LUNG & Veteran \\
        \midrule
         Cox & 0.83 & 0.56  & 0.60  \\
         
         RSF & 0.83 & 0.67 &  0.63 \\
         
         XGB & 0.87 & 0.67 &  0.75  \\
         
         DeepSurv & 0.50 & 0.50 &  0.52 \\
         
         DeepHit & 0.50 & 0.50 & 0.52  \\
        \bottomrule
    \end{tabular}
    \caption{c-index index for the models used to obtain the SurvLIME coefficients of Section \ref{sec:experiments}.}
    \label{tab:c_index}
\end{table}

For the UDCA dataset, Figure \ref{fig:Udca_weights} contains the feature importance for the models. It can be seen that  riskscore is the most important feature. The higher the value, the higher the CHF is for all the models, which is aligned with what is displayed in Table \ref{tab:udca_analysis}. For the Cox Proportional Hazards Model, the behaviour of the feature bili works in the opposite direction as it should be: according to Table \ref{tab:udca_analysis}, the higher the value of bili, the higher the risk of experiencing the event. However, according to Figure \ref{fig:Udca_weights}, the higher the value of bili, the lower the risk of experiencing the event. A possible explanation for this anomaly could be that bili feature correlates with riskscore feature, Pearson correlation coefficient between both of them is equal to 0.69. The Cox Proportional Hazards Model is very sensitive to this  phenomenon.

Out of all the models, the Cox Proportional Hazards Model is the only one whose coefficients can be directly compared with the SurvLIME's coefficients. Table \ref{tab:cox_udca} contains both sets of coefficients: the left column is for the coefficients of the Cox Proportional Hazards Model and the right column is for the median values of SurvLIME coefficients when it explains the Cox Proportional Hazards Model. Note that the median values are for the set $\{\bar{\boldsymbol\beta}_{1}^{s}, \dots, \bar{\boldsymbol\beta}_{n_t}^{s}\}$. Therefore, they are median values of mean values, since each vector $\bar{\boldsymbol\beta}_{j}^{s}$ is the mean vector across all the simulations. It can be seen that both sets of coefficients in Table \ref{tab:cox_udca} are close.

\begin{table}[!h]
    \centering
    \begin{tabular}{@{}l r r@{}}
        \toprule
        Feature & Cox & SurvLIME\\
        \midrule
        riskscore & 2.4397 & 1.6110\\
        stage & -0.0264 & -0.0392\\
        trt & -0.6480 & -0.3937\\
        bili & -1.7014 & -1.0954\\
        \bottomrule
    \end{tabular}
    \caption{Coefficients of Cox Proportional Hazards Model (middle column) and median values of SurvLIME coefficients (right column) for UDCA dataset.}
    \label{tab:cox_udca}
\end{table}

With regards to LUNG dataset, the feature importance is depicted in Figure \ref{fig:Lung_weights}. For  the Cox Proportional Hazards Model, the most important feature is ph.ecog. According to the model, the category that increases the most the CHF is 2 (ph.ecog\_2), followed by category 1 (ph.ecog\_1) and then by the category 0 (reference category). This is concordant with the values displayed in Table \ref{tab:lung_analysis}.

On the other hand, for the other two models, the most important one is age: the older an individual is, the higher the value of the CHF. The results shown in the Table \ref{tab:lung_analysis} are in the same direction: the older an individual is, the higher the probability of experiencing the event.

Table \ref{tab:cox_lung} contains the coefficients for the Cox Proportional Hazards Model and the median values of SurvLIME coefficients when it explains the Cox Proportional Hazards Model. The median values are calculated in the same way as they were calculated for the UDCA dataset. Note that  both sets of coefficients are close.

\begin{table}[!h]
    \centering
    \begin{tabular}{@{}l r r@{}}
        \toprule
        Feature & Cox & SurvLIME  \\
        \midrule
        ph.ecog\_2 & 0.6678 & 0.6117\\
        ph.ecog\_1 & 0.4419 & 0.4049\\
        age & 0.1551 &  0.1422\\
        ph.karno & 0.3206 &  0.2939\\
        pat.karno & -0.1805 & -0.1654\\
        wt.loss & -0.1491 & -0.1367\\
        sex & -0.2991 &  -0.2742\\
        \bottomrule
    \end{tabular}
    \caption{Coefficients of Cox Proportional Hazards Model (middle column) and median values of SurvLIME coefficients (right column) for LUNG dataset.}
    \label{tab:cox_lung}
\end{table}

Finally, Figure \ref{fig:Veterans_weights} shows the feature importance for each model. The three models consider that karno feature is the most important. According to the models, the higher the value of this feature, the lower the CHF is. This is aligned with what is shown in Table \ref{tab:veteran_analysis}. Table \ref{tab:cox_veteran} contains the coefficients for the Cox Proportional Hazards Model and the median values of SurvLIME coefficients when it explains this model. As for the UDCA as well as the LUNG datasets, both sets of coefficients are close.
\begin{table}
    \centering
    \begin{tabular}{@{}l r r@{}}
        \toprule
        Feature & Cox & SurvLIME  \\
        \midrule
         trt & 0.0979 & 0.0569\\
         prior & -0.0107 & -0.0138\\
         diagtime & -0.0166 & -0.0088\\
         age & -0.0454 & -0.0253\\
         celltype\_squamous & -0.5197 & -0.3690\\
         celltype\_smallcell & -0.0557 & -0.0461\\
         celltype\_large & -0.3110 & -0.2278\\
         karno & -0.7381 & -0.5251\\
        \bottomrule
    \end{tabular}
    \caption{Coefficients of Cox Proportional Hazards Model (middle column) and SurvLIME coefficients (right column) for Veteran dataset.}
    \label{tab:cox_veteran}
\end{table}

To conclude with this section, we have seen that our implementation captures the value of the coefficients when the machine learning model is the Cox Proportional Hazards Model.

\subsection{Simulated data and deep learning models}
\label{sec:simulated_data_dl}
As shown in Table \ref{tab:c_index}, DeepSurv and DeepHit did not perform better than a random model in any of the presented datasets. To show that our implementation of SurvLIME algorithm is able to obtain feature importance for deep learning models, we make use of simulated data. Concretely, the data generating process is the same as the one used for set 1 in Section \ref{sec:simulated_data}.

In order to train the deep learning models, we follow the same procedure as in Section \ref{sec:real_data}: $90\%$ of the individuals are used to train the models and $10\%$ are used to obtain the c-index as well as to obtain feature importance.

\begin{figure}
    \centering
    \begin{minipage}{0.49\textwidth}
        \includegraphics[width=1\textwidth]{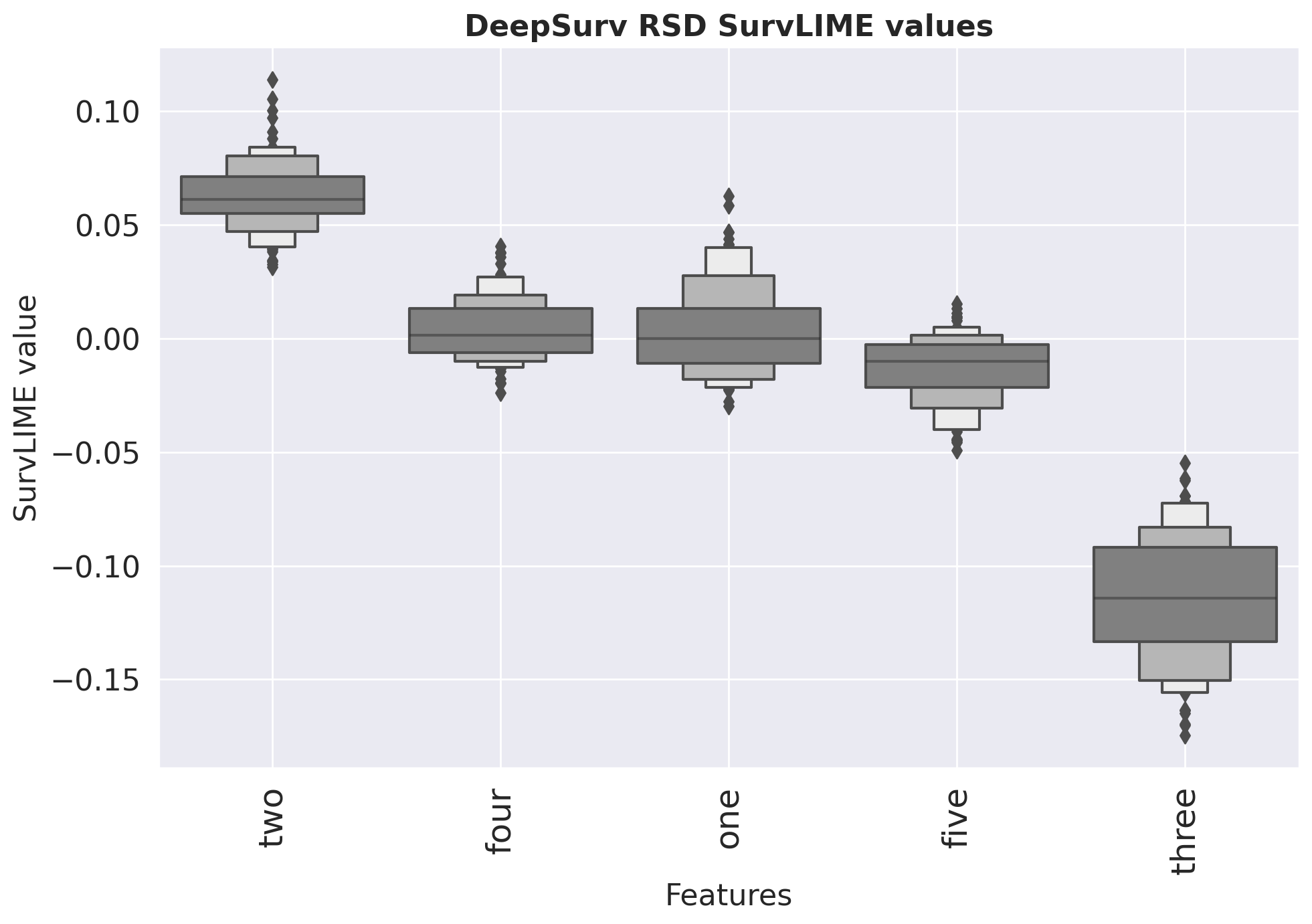}
    \end{minipage}
    \caption{Feature importance for the simulated data and DeepSurv model. The input parameter \code{with\_colour} is set to false.}
    \label{fig:dl_for_rsd}
\end{figure}

Table \ref{tab:c_index_dl_rsd} shows that both models have an acceptable predictive capacity on the simulated data. Using the same Monte-Carlo strategy, 100 different simulations are computed over the 100 test individuals. 
 The 100 mean values, $\{\bar{\boldsymbol\beta}_{1}^{s}, \dots, \bar{\boldsymbol\beta}_{100}^{s}\}$, computed across all the simulations are shown in Figure \ref{fig:dl_for_rsd}. It can be seen that the only features which deviate significantly from 0 are the feature two and three. This is aligned with the true coefficients, as shown in Table \ref{tab:dl_simulated}. In order to produce this table, we use the median values of the SurvLIME coefficients, i.e., the median across the set $\{\bar{\boldsymbol\beta}_{1}^{s}, \dots, \bar{\boldsymbol\beta}_{100}^{s}\}$. We omit to provide the SurvLIME coefficients for DeepHit since the values we have obtained are very similar to the values of DeepSurv.

\begin{table}[!ht]
    \centering
    \begin{tabular}{@{}l r@{}}
        \toprule
        Model & c-index \\
        \midrule
        DeepSurv &  0.70 \\
        
        DeepHit  &  0.68 \\
        \bottomrule
    \end{tabular}
    \caption{c-index for the deep learning models for the simulated data and DeepSurv model.}
    \label{tab:c_index_dl_rsd}
\end{table}

\begin{table}[!ht]
    \centering
    \begin{tabular}{@{}l r r@{}}
        \toprule
        Feature & Real coefficient & SurvLIME coefficient  \\
        \midrule
        two & 0.1 & 0.0711\\
        four & $10^{-6}$ & 0.0025  \\
        five & $10^{-6}$ &   -0.0088\\
        one & $10^{-6}$ &  -0.0045\\
        three & -0.15 & -0.1251 \\
        \bottomrule
    \end{tabular}
    \caption{Real coefficients used to generate the data (middle column) and median values of SurvLIME coefficients for DeepSurv model (right column).}
    \label{tab:dl_simulated}
\end{table}

\section{Conclusions}
\label{sec:conclusions}
In this paper \pkg{SurvLIMEpy} has been introduced in the form of a \proglang{Python} library. To the extent of our knowledge, this is the first module that tackles the problem of model explainability for time-to-event data in the \proglang{Python} programming language.

We have successfully demonstrated the validity of our implementation of the SurvLIME algorithm through a series of experiments with simulated and real datasets. Furthermore, we also grant flexibility to the algorithm by allowing users to adjust some of its internal parameters. 

Finally, a future research line would take into account how the feature importance evolves over time and incorporate it to \pkg{SurvLIMEpy}. Special care must be taken into account as the computational cost would increase significantly.

\section*{Acknowledgments}
This research was supported by the Spanish Research Agency (AEI) under projects PID2020-116294GB-I00 and PID2020-116907RB-I00 of the call MCIN/ AEI /10.13039/501100011033, the project 718/C/2019 funded by Fundació la Marato de TV3 and the grant 2020 FI SDUR 306 funded by AGAUR.

\bibliographystyle{unsrtnat}
\bibliography{survlime}

\newpage
\appendix
\section{Survival Analysis}
\label{appendix:surv-analysis}
Survival Analysis, also known as time-to-event analysis, is a branch of Statistics that studies the time until a particular event of interest occurs \citep{book_survival_2,Veterans_dataset}. It was initially developed in biomedical sciences and reliability engineering but, nowadays, it is used in a plethora of fields. 
A key point of a Survival Analysis approach is that each individual is represented by a triplet $(\mathbf{x}, \delta, \tau)$, where $\mathbf{x} = (x_1, x_2, \dots, x_p)\tr$ is the vector of features, $\tau$ indicates time to event or lost to follow-up time of the individual (it is assumed to be non-negative and continuous) and $\delta$ is the event indicator denoting whether the event of interest has been observed or not.

Given a dataset $\mathbf{D}$ consisting of $n$ triplets $(\mathbf{x_i}, \tau_i, \delta_i)$, $i \in \{ 1, \dots, n\}$, where $n$ is the number of individuals, Survival Analysis aims to build a model 
$\hat{H}  \colon \mathbb{R}^p \times \mathbb{R}^+ \to \mathbb{R}^+$, that allows to estimate the risk a certain individual $\mathbf{x}_*$ experiences the event at a certain time $t$. This risk estimator is given by $\hat{H}(\mathbf{x}, t)$.

\subsection{Censoring}
Censoring is a crucial phenomenon of Survival Analysis. It occurs when some information about individual survival time is available, but we do not know the exact survival time. It results in the event of interest not being observed for some individuals. This might happen when the event is not observed during the time window of the study, or the individual dropped out of the study by other uninterested causes. If this takes place, the individual $i$ is considered censored and $\delta_i = 0$. The three main types of censorship are:
\begin{itemize}
\item Right-censoring is said to occur when, despite continuous monitoring of the outcome event, the individual is lost to follow-up, or the event does not occur within the study duration \citep{prinja2010censoring}.
\item Left-censoring happens if an individual had been on risk for the event of interest for a period before entering the study.
\item Interval-censoring applies to individuals when the time until the event of interest is not known precisely (and instead, only is known to fall into a particular interval).
\end{itemize}

From the three of them, right-censoring, followed by interval-censoring, are the two most common types of censoring. Left-censoring is sometimes ignored since the starting point is defined by an event such as the entry of a individual into the study. % A visual representation for each type of censoring can be found in figure \ref{fig:censoring}.

If the event of interest is observed for individual $i$, $\delta_i = 1$ and $\tau_i$ correspond to the time from the beginning of the study to the event's occurrence respectively. This is also called an uncensored observation.

On the other hand, if the instance event is not observed or its time to event is greater than the observation window, $\tau_i$ corresponds to the time between the beginning of the study and the end of observation. In this case, the event indicator is $\delta_i = 0$, and the individual is considered to be censored.

% \citep{book_survival_1, book_survival_2}
\subsection{Survival Function}
The Survival Function is one of the main concepts in Survival Analysis, it represents the probability that the time to event is not earlier than time $t$ which is the same as the probability that a individual survives past time $t$ without the event happening. It is expressed as:
\begin{equation}
    S(t) = \Prob(T \geq t).
\end{equation}
It is a monotonically decreasing function whose initial value is 1 when $t = 0$, reflecting the fact that at the beginning of the study any observed individual is alive, their event is yet to occur. Its counterpart is the cumulative death distribution function $F(t)$ which states the probability that the event does occur earlier than time $t$, and it is defined as:
\begin{equation}
    F(t) = \Prob(T<t) = 1 -S(t).
\end{equation}
The death density function, $f(t)$, can also be computed as 
$f(t) = \frac{d}{dt} F(t) = - \frac{d}{dt} S(t).$

\subsection{Hazard Function and Cumulative Hazard Function} 
\label{hazard_function}
The second most common function in Survival Analysis is the Hazard Function or instantaneous death rate \citep{book_survival_1, book_survival_2}, denoted as $h(t)$, which indicates the rate of event at time $t$ given that it has not yet occurred before time $t$. It is also referred as risk score. It is also a non-negative function that can be expressed as:
\begin{equation}
    \begin{aligned}
        h(t) &= \lim_{\Delta t \to 0}\frac{\Prob(t \leq T \le t + \Delta t | T \geq t)}{\Delta t} \\
        &= \lim_{\Delta t \to 0}\frac{F(t+\Delta t) - F(t)}{\Delta t \cdot S(t)} = \frac{f(t)}{S(t)} = - \frac{\frac{d}{dt}S(t)}{S(t)}.
    \end{aligned}
\end{equation}

Similar to $S(t)$, $h(t)$ is a non-negative function but it is not constrained by monotonicity. Considering that $f(t) =-\frac{d}{dt}S(t)$, the Hazard Function can also be written as:
\begin{equation}
    h(t) = \frac{f(t)}{S(t)} = -\frac{d}{dt}\left[S(t)\right]\frac{1}{S(t)} = -\frac{d}{dt}[\ln S(t)].
    \label{hazard2}
\end{equation}

Integrating in both sides of Expression (\ref{hazard2}) from 0 to $t$ the Cumulative Hazard Function (CHF) is obtained and denoted as
$H(t) = \int_{0}^{t} h(r) \,dr$. It is related to the Survival Function by the following equation:
\begin{equation}
    S(t) = \exp(-H(t)).
    \label{SurvFunc}
\end{equation}

\subsection{Cox Proportional Hazards Model}
One of the historically most widely used semi-parametric algorithms for Survival Analysis is the Cox Proportional Hazards Model, published in \cite{CoxAlgorithm}. The model assumes a baseline Hazard Function $h_0(t)$ which only depends on the time, and a second hazard term $h(\mathbf{x}) = \exp(\hat{\boldsymbol\beta}\tr\mathbf{x})$ which only depends on the features of the individual. Thus, the Hazard Function $h(\mathbf{x}, t)$ in the Cox Proportional Hazards Model is given by:
\begin{equation}
    h(\mathbf{x}, t) = h_0(t)\exp(\hat{\boldsymbol\beta}\tr\mathbf{x}),
\end{equation}
where 
$\hat{\boldsymbol\beta} = (\hat{\beta}_1, \hat{\beta}_2, \dots, \hat{\beta}_p)\tr$ is the coefficient for the feature vector $\mathbf{x}$.

The Cox Proportional Hazards Model is a semi-parametric algorithm since the baseline Hazard Function $h_0(t)$ is unspecified. For two given individuals, their hazard's ratio is given by:
\begin{equation}
    \frac{h(\mathbf{x}_1, t)}{h(\mathbf{x}_2, t)} = \frac{h_0(t)\exp(\hat{\boldsymbol\beta}\tr\mathbf{x}_1)}{h_0(t)\exp(\hat{\boldsymbol\beta}\tr\mathbf{x}_2)} = \exp[\hat{\boldsymbol\beta}\tr(\mathbf{x}_1 - \mathbf{x}_2)].
\end{equation}
This implies that the hazard ratio is independent of $h_0(t)$. If it is then combined with Expression (\ref{SurvFunc}), the Survival Function can be computed as:
\begin{equation}
    S(\mathbf{x}, t) = \exp\left[-H_0(t)\exp(\hat{\boldsymbol\beta}\tr\mathbf{x})\right] = \left[S_0(t)\right]^{\exp(\hat{\boldsymbol\beta}\tr\mathbf{x})}.
\end{equation}

To estimate the coefficients $\hat{\boldsymbol\beta}$, Cox proposed a likelihood \citep{CoxAlgorithm} which depends only on the parameter of interest $\hat{\boldsymbol\beta}$. To compute this likelihood it is necessary to estimate the product of the probability of each individual that the event occurs at $\tau_i$ given their feature vector $\mathbf{x}_i$, for $i \in \{1, \dots, n\}$:
\begin{equation}
    \label{eq:likelihood}
    L(\boldsymbol\beta) = \prod_{i=1}^n\bigg [ \frac{\exp(\boldsymbol\beta\tr\mathbf{x}_i)}{\sum_{j \in R_i} \exp(\boldsymbol\beta\tr\mathbf{x}_j)}\bigg ]^{\delta_i},
\end{equation}
where $R_i$ is the set of individuals being at risk at time $\tau_i$. Note that $\hat{\boldsymbol\beta}$ is the vector that maximises Expression (\ref{eq:likelihood}), i.e,
$\hat{\boldsymbol\beta} = \argmax_{\boldsymbol\beta} L(\boldsymbol\beta)$.
%Note that if individual $i$ is censored, $\delta_i = 0$ and it does not contribute in the likelihood after its time-to-event.
\subsection{c-index}
The c-index, also known as concordance index \citep{c_index}, is a goodness-of-fit measure for time-dependant models. Given two random individuals, it accounts for the probability that the individual with the lower risk score will outlive the individual with the higher risk score.

In practical terms, given individuals $i$ and $j$ ($i \neq j$) as well as their risk scores, $H_i(t)$ and $H_j(t)$, and their times, $\tau_i$ and $\tau_j$, this probability is calculated taking into account the following scenarios:
\begin{itemize}
    \item If both are not censored, the pair $(i,j)$ is concordant if $\tau_i < \tau_j$ and $H_i(\tau_i) > H_j(\tau_i)$. 
    If $\tau_i > \tau_j$ and $H_i(\tau_j) > H_j(\tau_i)$, 
    the pair $(i,j)$ is discordant.
    \item If both are censored, the pair $(i,j)$ is not taken into account.
    \item For the remaining scenario, let suppose $i$ is not censored and $j$ is censored, i.e, $\delta_i=1$ and $\delta_j=0$. To make a decision, two scenarios are considered:
        \begin{itemize}
            \item If $\tau_j < \tau_i$, then the pair $(i,j)$ is not taken into account because $j$ could have experience the event if the experiment had lasted longer.
            \item If $\tau_i< \tau_j$, $i$ is the first individual whose event happens first (even if the experiment lasts longer). In this scenario, $(i,j)$ is concordant if $H_i(\tau_i) > H_j(\tau_i)$. Otherwise, this pair is discordant.
        \end{itemize}
\end{itemize}

Once all the scenarios are taken into account and considering all pair $(i,j)$ such that $i \neq j$, the c-index can be expressed as
\begin{equation}
    \label{eq:c_index}
    \frac{\text{concordant pairs}}{\text{concordant pairs} + \text{discordant pairs}}.
\end{equation}
The more concordant pairs, the better the model is estimating the risk. Therefore, the higher the c-index, the more accurate the model is. The maximum value for the c-index is 1. A value equal to 0.5 (or lower) means that the models performs as a random model. More details are provided in \cite{c_index} or in \cite{c_index_more}.

\end{document}